\documentclass{egpubl}
\Paper
\usepackage[T1]{fontenc}
\usepackage{dfadobe}  
\usepackage{cite}
\BibtexOrBiblatex
\electronicVersion
\PrintedOrElectronic
\ifpdf \usepackage[pdftex]{graphicx} \pdfcompresslevel=9
\else \usepackage[dvips]{graphicx} \fi
\usepackage{egweblnk} 
\usepackage{microtype}
\usepackage{amsfonts}
\usepackage{nicefrac}
\usepackage{enumitem}
\usepackage{booktabs}
\usepackage{mathtools}
\usepackage{placeins}
\usepackage[capitalize]{cleveref}
\Crefname{section}{Section}{Sections}
\crefname{section}{Sec.}{Secs.}
\Crefname{table}{Table}{Tables}
\crefname{table}{Tab.}{Tabs.}
\Crefname{figure}{Figure}{Figures}
\crefname{figure}{Fig.}{Figs.}
\usepackage{multirow}
\usepackage{multicol}
\usepackage[table]{xcolor}
\definecolor{1st}{RGB}{102,194,164} %
\definecolor{2nd}{RGB}{178,226,226} %
\definecolor{3rd}{RGB}{217,248,241} %
\definecolor{1stText}{RGB}{57,146,116} %

\usepackage{algorithmic}
\usepackage{algorithm}

\newcolumntype{C}[1]{>{\centering\arraybackslash}p{#1}}
\title{Make It Flow: Animating Water in 3D Gaussian Splatting Scenes\vspace{-20pt}}

\author[K. Keune \& F. Hahlbohm et al.]
{\parbox{\textwidth}{\centering%
Kjell Keune\hspace{2pt}\orcid{0009-0008-8233-5678}\hspace{60pt}
Florian Hahlbohm\hspace{2pt}\orcid{0009-0004-8710-1433}\\
Jannis Möller\hspace{2pt}\orcid{0000-0003-0870-614X}\hspace{10pt}
Susana Castillo\hspace{2pt}\orcid{0000-0003-1245-4758}\hspace{10pt}
Martin Eisemann\hspace{2pt}\orcid{0000-0002-8673-4405}\hspace{10pt}
Marcus Magnor\hspace{2pt}\orcid{0000-0003-0579-480X}
}\\
\parbox{\textwidth}{\centering%
Computer Graphics Lab, TU Braunschweig, Germany\hspace{7pt}\texttt{\{lastname\}@cg.cs.tu-bs.de}\\
\vspace{15pt}
\large{Overview video at \url{https://youtu.be/qf04D73kgrc}}
}
}

\begin{document}

\teaser{
\vspace{-20pt}
\centering
\includegraphics[width=0.9\textwidth]{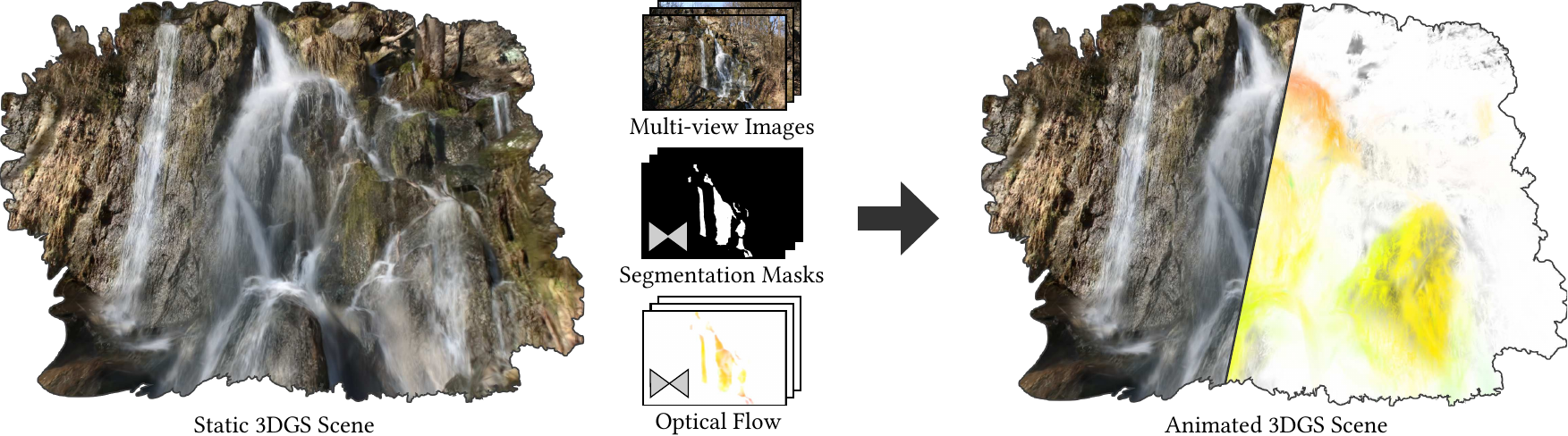}
\caption{%
We generate seamlessly looping water animations in static 3D Gaussian Splatting scenes via a cinemagraph-inspired approach that lifts optical flow into a coherent 3D Eulerian motion field. This field is iteratively applied to Gaussians via an unscented transform, producing temporally coherent fluid motion and enabling real-time rendering and immersive free-viewpoint exploration of dynamically animated scenes with minimal overhead as evident in our video.
}\label{fig:teaser}
}

\maketitle
\begin{abstract}
Explicit 3D scene representations such as 3D Gaussian Splatting (3DGS) have recently enabled high-quality, real-time novel view synthesis.
However, these methods assume static scenes, limiting realism in the presence of dynamic phenomena ubiquitous in real-world environments.
While coherent motion (e.g., humans or vehicles) can be captured and reconstructed, high-frequency effects such as flowing water remain intractable under casual capture conditions, where dense multi-view, high-speed acquisition is not feasible.
We present a method to animate water in static 3DGS scenes, enabling visually plausible motion without requiring dynamic capture or physical simulation.
Our approach estimates 2D optical flow from reference images or model renderings and lifts it into a consistent 3D Eulerian motion field.
We propose a cinemagraph-inspired algorithm that applies this motion field to deform Gaussian primitives, producing temporally seamless looping animations.
The resulting representation retains real-time rendering performance, significantly enhances perceptual realism, and preserves free-viewpoint navigation within the scene.
We evaluate our method on a newly captured dataset of in-the-wild waterfall scenes and demonstrate clear improvements over prior work in dynamic Gaussian splatting and 3D cinemagraph generation in visual quality and computational efficiency.
\end{abstract}

\section{Introduction}
\label{sec:introduction}
Novel view synthesis has significantly advanced the reconstruction and interactive exploration of real-world environments.
Recent progress in 3D Gaussian Splatting (3DGS)~\cite{kerbl3Dgaussians} combines the visual fidelity of neural radiance field approaches~\cite{mildenhall2020nerf} with the efficiency of point-based rendering, enabling high-quality real-time rendering even on resource-constrained platforms.
Despite these advances, reconstructed scenes remain predominantly static and often fail to capture dynamic phenomena that are essential for visual realism.
This limitation is particularly evident in fluid regions, where 3D Gaussian primitives frequently produce temporally inconsistent or structurally unstable reconstructions.
As a result, phenomena such as flowing water often appear unnaturally frozen.
In this work, we show that introducing plausible motion specifically in these regions can substantially enhance the perceived realism and liveliness of reconstructed scenes.

Recent dynamic Gaussian Splatting methods have made substantial progress toward extending static reconstruction frameworks to dynamic scenes, but primarily focus on explicitly moving foreground objects~\cite{luiten2023dynamic, wu20244dgs, yang20244dgaussians}.
In contrast, subtle continuous background phenomena such as rivers or waterfalls remain largely underexplored.
Moreover, many dynamic reconstruction approaches rely on multi-view synchronized capture setups, limiting their applicability to casually captured or in-the-wild data.
An alternative direction augments static 3DGS reconstructions with synthesized dynamics.
While physics-based methods~\cite{xie2024physgaussian} can produce convincing simulations across materials including fluids, they incur substantial computational cost and are difficult to apply in real-time or casual capture settings.
Instead of targeting physically accurate simulation, which is often unnecessary for visual realism, we aim to generate visually plausible motion that improves scene perception while remaining efficient.

To this end, we draw inspiration from cinemagraphs, which introduce temporally looping motion into otherwise static imagery and are particularly effective for fluid phenomena in 2D~\cite{holynski2021animating}.
Prior work extending cinemagraphs to 3D either combines layered depth representations with neural networks~\cite{li2023_3dcinemagraphy}, resulting in slow rendering and limited free-viewpoint capability, or relies on self-similarity in Gaussian point clouds~\cite{li2024loopgaussian}, which provides insufficient structure to recover realistic fluid motion.
These limitations motivate a more efficient approach for visually plausible water animation in static 3DGS scenes.

Given a 3DGS model and its training views (or alternatively model renderings), we estimate per-view fluid flow with an existing pretrained network~\cite{mahapatra2021controllable}.
Our first contribution is lifting these multi-view estimates into a grid-based 3D Eulerian motion field via depth-based reprojection, refined via static/dynamic voxel hit ratios and view-hit thresholding.
Second, we adopt the concept of symmetric warping from 2D cinemagraphs~\cite{holynski2021animating}, which is defined for pixels, and reformulate it for 3D Gaussians treated as Lagrangian particles: volume-aware covariance propagation, density-preserving opacity blending, and randomized per-Gaussian temporal offsets replace the feature warping and inpainting used in 2D.
The resulting method integrates directly into the 3DGS rendering pipeline, enabling real-time animation of static scenes with low overhead and allowing interactive exploration of dynamically enhanced environments (cf.\ \cref{fig:teaser}).

For evaluation, we introduce a new dataset of seven casually captured 3DGS scenes containing fluid phenomena such as waterfalls.
We compare against dynamic Gaussian Splatting~\cite{yang20244dgaussians} and recent 3D cinemagraph methods~\cite{li2023_3dcinemagraphy, li2024loopgaussian} qualitatively and in terms of efficiency.
As evident in our video, our method produces more realistic and temporally coherent novel view renderings while remaining computationally efficient.

In summary, we make the following contributions:
\begin{itemize}
\item The first method to animate fluid regions in a static 3DGS scene, without dynamic capture, retraining, or simulation.
\item An efficient, 3DGS-based approach for lifting multi-view optical flow into a 3D Eulerian motion field.
\item A deformation framework for temporally looping, visually plausible fluid animation, integrated into a real-time 3DGS rendering system for interactive exploration of animated scenes.
\item A curated dataset of seven casual scenes with fluid motion.
\end{itemize}

We will release our full implementation, including preprocessing scripts, interactive GUI, and dataset.

\section{Related Work}
\label{sec:related_work}
Our work extends novel view synthesis by integrating cinemagraph-inspired motion into static 3D scene representations, with a focus on efficient and perceptually plausible fluid animation.

\subsection{Novel View Synthesis}
Early image-based rendering and light field methods~\cite{debevec1996mar, gortler1996lumigraph, levoy1996lightfields} are limited by sparse viewpoint coverage.
Neural radiance fields (NeRF)~\cite{mildenhall2020nerf} significantly advanced novel view synthesis, but are expensive to train and render, which motivated more efficient scene parameterizations~\cite{yu2021plenoxels, sun2022dvgo, mueller2022instant, Chen2022tensorf}.
3D Gaussian Splatting (3DGS)~\cite{kerbl3Dgaussians} models scenes as anisotropic Gaussians optimized for real-time rasterization and forms the basis of our approach.

Extensions of both NeRF and 3DGS to dynamic scenes introduce temporal variation either through explicit time-conditioned representations or learned deformation fields~\cite{pumarola2021dnerf, park2021nerfies, cao2023hexplane, fridovichkeil2023kplanes, fang2022tineuvox, lin2023gaussianflow, yang20244dgaussians, wu20244dgs}.
These methods achieve compelling results for structured and well-observed motion such as humans or rigid objects, but typically assume controlled multi-view capture setups or availability of strong data-driven priors.
Such assumptions restrict their applicability to casually captured or in-the-wild scenes.
Moreover, they are less suitable for subtle, continuous background phenomena such as water, where motion is locally smooth, temporally repetitive, and driven more by perceptual consistency than strict geometric reconstruction.
\begin{figure*}[t]
\centering
\includegraphics[width=1.0\linewidth]{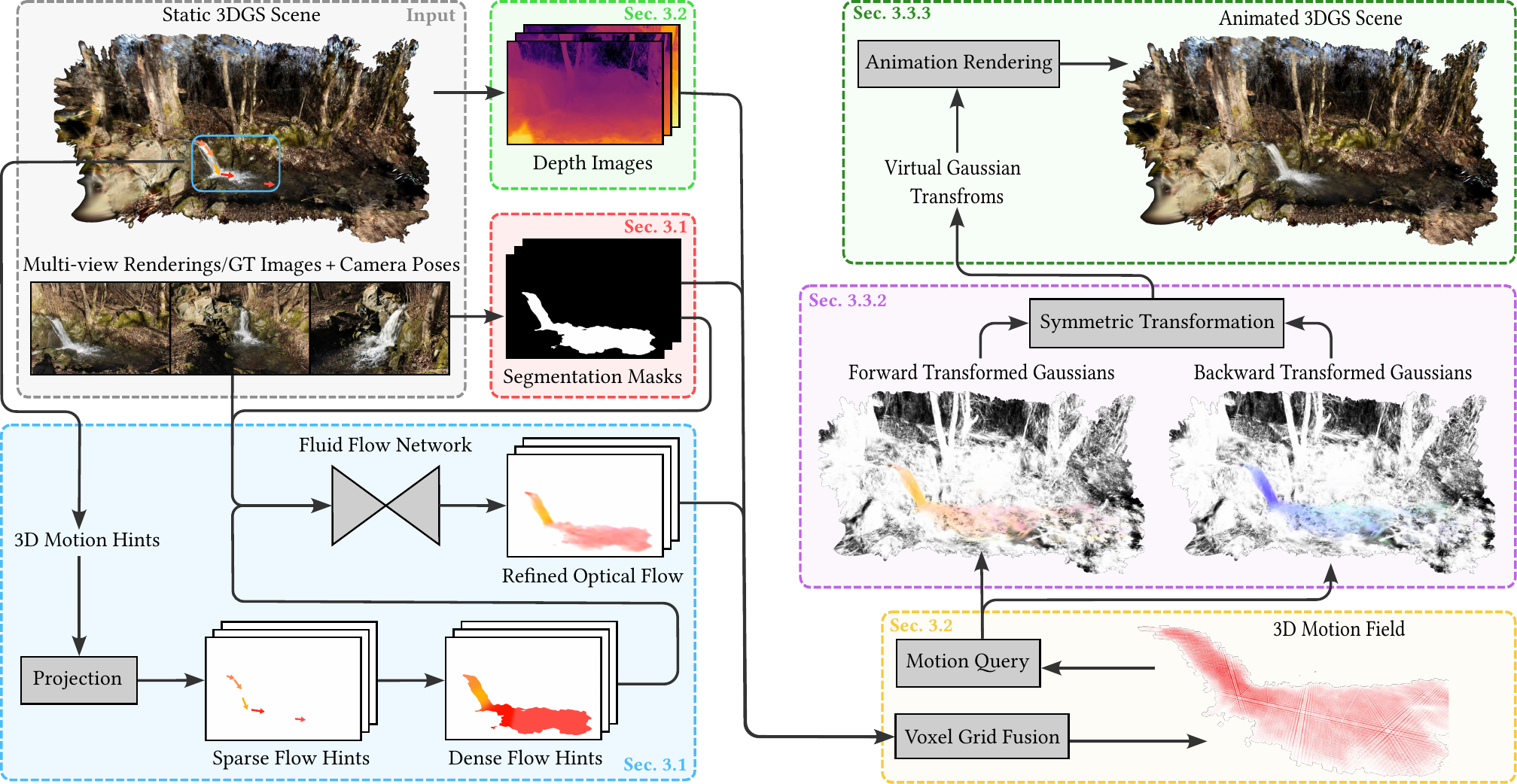}
\caption{%
We generate optical flow depicting fluid motion from multi-view observations of a static 3DGS reconstruction by refining user-defined motion hints with segmentation masks and GAN-based flow estimation. Utilizing depth renderings from the 3DGS model, we reproject optical flow into a Eulerian motion field and create seamlessly looping animations by transforming multiple instances of dynamic Gaussians using a symmetric animation technique.
}\label{fig:pipeline}
\end{figure*}

\subsection{Cinemagraphs and Motion Synthesis}
Cinemagraphs and the closely related video textures~\cite{schodl2000vidtex} introduce looping motion into static imagery while preserving spatial consistency.
Early work focused on manual or semi-automatic creation, followed by numerous methods for automatic loop synthesis and fluid animation in videos~\cite{liao2013automatedvideolooping, liao2015fastseamlessloops, tompkin2011towardsmomentimagery}.
These techniques have demonstrated that fluid regions such as waterfalls are particularly well-suited for perceptually convincing motion synthesis.

Recent learning-based approaches extend cinemagraph generation to single-image or sparse-input settings using generative models or optical flow prediction~\cite{hao2018vidgen, gao2018im2flow, endo2019animatinglandscape}.
More recently, diffusion-based image-to-video models have enabled controllable cinemagraph synthesis with localized motion editing and user-guided spatial constraints~\cite{le2026cinemagraphvideodiffusion}.
While visually compelling, these approaches remain computationally expensive and typically require repeated inference for long-duration or seamlessly looping animations, limiting interactivity and real-time applicability.

To address this, recent works learn static or quasi-static motion fields that enable efficient looping animations~\cite{holynski2021animating, mahapatra2021controllable, halperin2021loops}.
These approaches are effective for synthesizing fluid motion and form the foundation of our motion generation framework.

Recent work has further explored lifting cinemagraph synthesis into 3D scene representations.
Approaches based on layered depth images or multi-view video synthesis project 2D motion into 3D representations, but remain limited in viewpoint flexibility or reconstruction quality~\cite{li2023_3dcinemagraphy, ma2023videoloop}.
More recent methods leverage neural or Gaussian-based scene representations to synthesize motion directly in 3D~\cite{li2024loopgaussian, jia2025physic}.
However, these methods often rely on appearance-based similarity, specialized motion priors, or computationally expensive learned pipelines, which work well for their specific application but limit their ability to model coherent directional fluid motion in unconstrained scenes.

Related approaches also augment static Gaussian scenes with specialized scene dynamics such as snowfall using physics-guided diffusion models~\cite{fiebelman2026letsnow}, or fluids using physics-based simulation with artificially added fluid particles rather than animating reconstructed water~\cite{feng2025splashing}.
Physics-based methods incur substantial computational cost, require careful per-scene setup, or noticeably alter scene appearance.
Recent methods targeting oscillatory scene dynamics, such as swaying vegetation, employ specifically tailored motion models~\cite{shih2024ambientdynamics, luo2026growflow, tzathas2026oscilatorymotion}.
While effective for such structured oscillatory motion, these assumptions do not readily generalize to fluid phenomena.

In contrast to dynamic reconstruction, learned 3D cinemagraph approaches and physics-based simulation, we augment static 3DGS scenes using motion derived from 2D flow priors, enabling efficient and controllable fluid animation without reconstructing a fully dynamic scene.

\section{Method}
\label{sec:method}
We propose an end-to-end pipeline for animating fluid regions in a static 3D Gaussian Splatting~(3DGS)~\cite{kerbl3Dgaussians} reconstruction (cf.\ \cref{fig:pipeline}).
In contrast to previous methods~\cite{li2023_3dcinemagraphy, ma2023videoloop}, our objective is a complete animated 3D scene observable from arbitrary viewing directions.
Besides its real-time rendering capabilities, 3DGS offers an explicit scene representation based on a set of points $\mathcal{G}$ with anisotropic Gaussian extent that are straightforward to animate.
Specifically, each Gaussian is described by its mean $\mathbf{\mu}\in\mathbb{R}^3$, 3D covariance $\Sigma\in\mathbb{R}^{3\times3}$, opacity $o\in[0,1]$, and view-dependent color $c\in\mathbb{R}^3$ computed from Spherical Harmonics coefficients $\in\mathbb{R}^{48}$.
This representation is then optimized from multi-view RGB images $\mathcal{I}_{1,\cdots,N}$ with known camera poses.
To render an image, 3DGS linearly approximates the perspective projection onto the image plane at a point $\mathbf{x}$ (usually $\mathbf{\mu}$)~\cite{zwicker2001ewasplat}:
\begin{equation}\label{eq:jacobian}
J_{n,\,\mathbf{x}} = \begin{pmatrix}
\frac{f}{\mathbf{\hat{x}}_3} & 0 & -\frac{f \mathbf{\hat{x}}_1}{(\mathbf{\hat{x}}_3)^2}\\
0 &\frac{f}{\mathbf{\hat{x}}_3}& -\frac{f \mathbf{\hat{x}}_2}{(\mathbf{\hat{x}}_3)^2}
\end{pmatrix} R_n, \; \text{with} \; \mathbf{\hat{x}} = R_n \mathbf{x} + \mathbf{t}_n
\end{equation}
where $\mathbf{\hat{x}}_i$ denotes the $i$-th component of $\mathbf{\hat{x}}$, $(R_n, \mathbf{t}_n)$ the world-to-camera transformation of the $n$-th view, and $f$ the focal length.
Optionally, $\mathcal{I}_{1,\cdots,N}$ can be extended by renderings from user defined camera poses.
For each pixel $\mathbf{p}$ in a rendered image $\mathcal{I}_n$, color is accumulated by alpha blending Gaussian contributions:
\begin{equation}\label{eq:alpha_blending}
\mathcal{I}_n(\mathbf{p}) = \sum_{i=1}^{M} c_i T_i o_i g_i(\mathbf{p}) \quad \text{with} \quad T_i = \prod_{j=1}^{i-1} (1 - o_j g_j(\mathbf{p})),
\end{equation}
where $g_i(\mathbf{p})$ is the $i$-th Gaussian value at the pixel $\mathbf{p}$.
Since our method assumes $\mathcal{G}$ and $\mathcal{I}_{1,\cdots,N}$ are provided as input, no training modifications and minimal renderer changes (see \cref{sec:rendering}) are required.
In the following subsections, we describe each step in our pipeline for animating fluids in such 3DGS models.

\subsection{Optical Flow Estimation}
\label{sec:motion_flow}

To construct a 3D motion field, we first estimate 2D optical flow $\mathcal{F}_{1,\cdots,N}$ for each input view.
Since the input images are static and non-time-consecutive, conventional optical flow methods~\cite{teed2020raft} are not applicable.
Instead, following prior cinemagraph approaches~\cite{holynski2021animating, mahapatra2021controllable}, we employ a specialized neural network trained on fluid motion.
For improved controllability and cross-view consistency, the network is guided by user-defined motion hints and segmentation masks.

We define a set of global 3D motion hints $\mathcal{H} = \left\{(\mathbf{p}_l,\mathbf{d}_l)\right\}_{l=1}^L$ interactively within the 3DGS scene using our GUI.
Each hint consists of a 3D position $\mathbf{p}_l\in\mathbb{R}^3$ and a motion direction $\mathbf{d}_l\in\mathbb{R}^3$, where $\lVert \mathbf{d}_l \rVert_2$ controls motion strength.
Defining hints directly in 3D ensures consistent motion guidance across views.
The hints are projected into each camera view using the projection operator $\pi_n$, resulting in sparse 2D flow hints $\mathcal{H}_n = \left\{(\pi_n(\mathbf{p}), \pi_n(\mathbf{p + d}) - \pi_n(\mathbf{p})) \mid (\mathbf{p},\mathbf{d})\in\mathcal{H}\right\}$ that are converted to sparse flow maps:
\begin{equation}
    \mathcal{F}^{\text{sparse}}_n(\mathbf{x})=
    \begin{cases}
    \frac{\mathbf{d}}{r^\mathcal{H}}, & \text{if } \exists\,\mathbf{d} \text{ with }(\mathbf{x},\mathbf{d})\in\mathcal{H}_n\\
    0, & \text{otherwise}
    \end{cases} \,,
\end{equation}
with a regulation factor $r^\mathcal{H}=\operatorname{median}\left\{ \lVert \mathbf{d} \rVert_2 \mid (\mathbf{p},\mathbf{d})\in\mathcal{H}_n \right\}$ to limit the extent of projected motion hints in screen space.

To restrict motion to fluid regions, we additionally compute binary segmentation masks $\mathcal{S}_{1,\cdots,N}$ for all input images using the promptable segmentation model SAM3~\cite{carion2026sam3}.
These masks are used both to guide the motion synthesis network and to constrain subsequent 3D motion field fusion.
In practice, simple prompts such as ``water'' or ``waterfall'' are sufficient for robust segmentation in most scenes.
For cases requiring additional user control, we use an interactive segmentation tool~\cite{cheng2024cutie}.

Following Mahapatra and Kulkarni~\cite{mahapatra2021controllable}, the sparse flow maps are converted into dense flow maps $\mathcal{F}^{\text{dense}}_{1,\cdots,N}$ via distance-weighted interpolation within the segmented regions (see \cref{app:dense}).
Finally, a learned flow refinement network takes RGB images $\mathcal{I}_{1,\cdots,N}$, segmentation masks $\mathcal{S}_{1,\cdots,N}$, and dense flow maps $\mathcal{F}^{\text{dense}}_{1,\cdots,N}$ as input to generate high-quality fluid flow maps $\mathcal{F}_{1,\cdots,N}$.
We adopt the SPADE-based U-Net architecture of Mahapatra and Kulkarni~\cite{mahapatra2021controllable} using the implementation and pretrained weights provided by Li~et~al.~\cite{li2023_3dcinemagraphy}.

\subsection{3D Motion Field}\label{sec:motion}

In contrast to motion estimated from Gaussian self-similarity~\cite{li2024loopgaussian}, which suits subtle motion such as swaying vegetation but not dense directional fluid flow, or to flow projected onto a single-view layered depth image~\cite{li2023_3dcinemagraphy}, we project multi-view optical flow into the 3DGS scene to obtain a coherent 3D motion field $\mathcal{M}$.

\paragraph*{Depth Rendering.}\label{sec:motion_depth}
In order to reproject 2D optical flow into the scene, depth information is required.
To this end, we render depth maps $\mathcal{D}_{1,\cdots,N}$ from the 3DGS model for all input views.
We compute the depth of each pixel via alpha blending as defined in \cref{eq:alpha_blending}, where the color of each Gaussian is replaced by its depth $\mathbf{\hat{\mu}}_3$ (cf.\ \cref{eq:jacobian}).
Note that we divide by the final alpha value of the pixel to obtain meaningful depth for semi-transparent pixels.
We further highlight that using depth renderings of the underlying 3DGS model aligns the motion field with Gaussians that can be animated.

\paragraph*{Multi-View Reprojection.}\label{sec:motion_mvr}
Our setup for the motion field fusion can be thought of as a multi-view reprojection problem~\cite{hartley2003mvcv, torralba2024foundcv}, with $n$ known projections $v_n = \mathcal{F}_n(x_n)\in\mathbb{R}^2$ of an unknown vector $\mathbf{v}\in\mathbb{R}^3$.
This vector is anchored at $\mathbf{x} = \pi^{-1}_n(x_n, \mathcal{D}_n(x_n))$, which is the common point when unprojecting each $x_n$ into world space using depth information.
The goal is minimizing the geometric non-linear reprojection error:
\begin{equation}
 \displaystyle\min_{\mathbf{v}}\displaystyle\sum_{n}\left\lVert \pi_n(\mathbf{x} + \mathbf{v}) - (x_n + v_n) \right\rVert_2^2 \,.
\end{equation}
As $v_n$, i.e., the optical flow is usually small, it is reasonable to linearly approximate the perspective projection by first-order Taylor expansion $\pi_n(\mathbf{x} + \mathbf{v}) \approx \pi_n(\mathbf{x}) + J_{n,\,\mathbf{x}}\mathbf{v}$, with $J_{n,\,\mathbf{x}}$ as defined in \cref{eq:jacobian}.
Since $\pi_n(\mathbf{x}) = x_n$ is known, we can solve a simplified linear system of equations for $\mathbf{v}$ with the iterative Gauss-Newton algorithm:
\begin{equation}\label{equ:reprojection_linsys}
\underbrace{\left(\sum_n J_{n,\mathbf{x}}^\top J_{n,\mathbf{x}}\right)}_\mathbf{A}
\mathbf{v}
=
\underbrace{\sum_n J_{n,\mathbf{x}}^\top v_n}_\mathbf{b}.
\end{equation}

\subsubsection{Voxel Grid Fusion}\label{sec:motion_fusion}
We implement the motion field $\mathcal{M}: \mathbb{R}^3\rightarrow\mathbb{R}^3$ as a regular voxel grid with fixed voxel size and store motion at voxel centers $\mathbf{c}\in C$.
To enable querying for motion $\mathbf{v}$ at an arbitrary scene point $\mathbf{x}$, we apply trilinear interpolation of the neighboring voxel centers of $\mathbf{x}$.
The trilinear interpolation weights $w^{\text{tri}}$ define the motion contribution of each involved $\mathbf{c}$, based on distance of $\mathbf{c}$ and $\mathbf{x}$.
Every query for an $\mathbf{x}$ outside the bounding box of the grid results in $\mathbf{v} = \vec{0}$.
The bounding box is initialized by the dimensions of the point cloud containing all reprojected dynamic pixels.

For fusion, we apply the concepts of multi-view reprojection and accumulate $\mathbf{A}$ and $\mathbf{b}$ (\cref{equ:reprojection_linsys}), per voxel over all views and solve for $\mathbf{v}$ afterward.
We leverage GPU-accelerated linear algebra to efficiently solve the Gauss–Newton system in parallel across voxels.
After fusion, $\mathcal{M}$ is converted into a sparse voxel grid, for memory efficient storage on disk.
Note that fusion is only performed once, since we use $\mathcal{M}$ as a static velocity field for all animation frames similar to Holynski~et~al.~\cite{holynski2021animating}.

In a first pass through the input data we approximate $\mathbf{v}$ at $\mathbf{c}\in C$ for all voxels based on $\mathcal{F}_n$ and $\mathcal{D}_n$ of the projected views $n$.
Pixels within dynamic regions $x\in\mathcal{S}_n$ are reprojected with $\mathbf{x} = \pi^{-1}_n(x,\, \mathcal{D}_n(x))$, and grouped to the corresponding voxels.
Grouping uses either hard indexing within the voxel bounds (each pixel has weight $1$) or trilinear splatting to neighboring voxels (weight $w^{\text{tri}}$).
Per view and voxel, the weighted flow values are averaged to obtain $v_n$, and $w_n$ denotes the sum of weights.
With the Jacobian $J_{n,\,\mathbf{c}}$ evaluated at the voxel center, $\mathbf{A}_{\mathbf{c}}$ and $\mathbf{b}_{\mathbf{c}}$ are accumulated over all views, each view contributing once per voxel.
After accumulating all views, the per-voxel 3\texttimes3 linear systems $\mathbf{A}_{\mathbf{c}}\mathbf{v} = \mathbf{b}_{\mathbf{c}}$ are solved and motion is stored such that $\mathcal{M}(\mathbf{c}) = \mathbf{v}$.
For refinement, we also accumulate dynamic pixel hits $w^{\text{dyn}}_{\mathbf{c}} = \sum_n w_n$ and view hits $w^{\text{view}}_{\mathbf{c}}$, i.e., how many views contributed motion to a voxel.

\subsubsection{Refinement}
Wrong depth information from misplaced Gaussians causing wrong pixel reprojection, or voxels intersecting the boundary of static and dynamic regions, can lead to motion within static regions.
To this end, we refine $\mathcal{M}$ by deleting these incorrect voxels by performing a second pass through the input data.
Pixels outside the masked region $\mathcal{S}_n(x)=0$ are projected and grouped by voxels as before, but rather than performing reprojection, now only static hits are tracked as $w^{\text{stat}}_{\mathbf{c}}$ weighted by $w^{\text{tri}}$ if necessary.
Intuitively, a voxel observed more often as static than as dynamic most likely lies in a static region, and its motion is reduced accordingly:
\begin{equation}
   \mathcal{M}(\mathbf{c}) =\left(1-\operatorname{min}\left(1,\,\frac{w^{\text{stat}}_{\mathbf{c}}}{w^{\text{dyn}}_{\mathbf{c}}} \right)\right)\cdot\mathbf{v} \,.
\end{equation}

We also find that the multi-view reprojection problem becomes ill-posed when there are limited observations, especially if the associated viewing directions are mostly collinear.
In practice, we therefore assume that a number of observations below a threshold is insufficient to reconstruct a plausible $\mathbf{v}$.
Consequently, we set the motion of voxels with view hits $w^{\text{view}}_{\mathbf{c}}$ below this threshold $r^{\text{view}}$ to zero, further reducing the size of $\mathcal{M}$.

We show pseudocode for our 3D motion field fusion approach and refinement techniques in \cref{app:motion_field_creation}.

\subsection{Gaussian Animation and Rendering}
\label{sec:animation}
We treat 3D Gaussians as Lagrangian particles that undergo deformation within Eulerian grids.
The majority of Gaussians reside in static scene regions and will therefore not be deformed by $\mathcal{M}$.
This enables us to efficiently segment the scene into static $\mathcal{G}^{\text{stat}} = \mathcal{G}\setminus\mathcal{G}^{\text{dyn}}$ and dynamic Gaussians $\mathcal{G}^{\text{dyn}}=\{ g\in\mathcal{G} \mid \lVert\mathcal{M}(\mu)\rVert_2 > 0\}$ by querying motion at their means.

\subsubsection{Transforming Gaussians}\label{sec:animation_ut}
To achieve volume-aware transformation of 3D Gaussians, we iteratively apply \textit{Unscented Transform}~\cite{julier1997ut} with velocities queried from $\mathcal{M}$ to transform mean and covariance of each $g\in\mathcal{G}^{\text{dyn}}$.
The volume of a 3D Gaussian is approximated by seven sigma points:
\begin{equation}\label{eq:sigma_points}
    \sigma_i(\mu,\Sigma) =
    \begin{cases}
    \mu & i=0 \\
    \mu + (-1)^i \sqrt{(3+\lambda)}\sqrt{\Sigma}_{\lceil{i/2}\rceil} & i=1,\dots,6
    \end{cases} \,,
\end{equation}
with $\sqrt{\Sigma}_i$ being the $i$-th column of the Cholesky-decomposed covariance and $\lambda\!=\!3\alpha^2\!-\!3$ and $\alpha$ a parameter to control the spread around $\mu$.
The sigma points have corresponding weights for mean and covariance defined as
\begin{align}
w^\mu_i =
\begin{cases}
\frac{\lambda}{3+\lambda}&i=0 \\
\frac{1}{2(3+\lambda)}&i=1,\dots,6 
\end{cases}, \;
w^\Sigma_i = w^\mu_i +
\begin{cases}
3-\alpha^2 & i=0 \\
0 & i=1,\dots,6 
\end{cases}.
\end{align}
The Unscented Transform can be applied iteratively via Euler integration to acquire updated mean $\mu_{k+1}$ and covariance $\Sigma_{k+1}$ for any Gaussian $g$ and iteration $k$.
The sigma points are displaced by the motion queried from $\mathcal{M}$ as
\begin{equation}
    \tilde{\sigma}_i = \sigma_i(\mu_k,\Sigma_k) + r^\mathcal{M}\mathcal{M}\left(\sigma_i(\mu_k,\Sigma_k)\right) \,.
\end{equation}
The recursion is initialized with the original Gaussian parameters $\mu_0 = \mu, \, \Sigma_0=\Sigma$.
We also apply a regulation factor $r^\mathcal{M}$ to manually control the motion extent in cases where $\mathcal{M}$ does not contain physically meaningful velocities.
The displaced sigma points $\tilde{\sigma}_i$ are used to update mean and covariance as
\begin{equation}\label{eq:iterative_ut}
\mu_{k+1} = \sum^6_{i=0} w^\mu_i \tilde{\sigma}_i \;, \quad \Sigma_{k+1} = \sum^6_{i=0} w^\Sigma_i \left(\tilde{\sigma}_i - \mu_{k+1} \right) \left(\tilde{\sigma}_i - \mu_{k+1} \right)^\top.
\end{equation}

\subsubsection{Symmetric Animation}
\label{sec:animation_symmetric}
Naturally, $\mathcal{M}$ contains motion sources, i.e., regions from which particles move away without new particles moving in, such as the top of a waterfall.
For a looping animation, we extend the symmetric warping technique of Holynski~et~al.~\cite{holynski2021animating} to 3D Gaussians by applying forward and backward transformations.
Each $g\in\mathcal{G}^{\text{dyn}}$ is forward transformed through $K$ iterations of \cref{eq:iterative_ut} into a keyframe-sequence $\mathcal{A}^{\text{fw}} = \big( (\mu_k, \Sigma_k) \big)^K_{k=0}$.
Analogously, a backward keyframe-sequence $\mathcal{A}^{\text{bw}} = \big( (\mu_{K-k}, \Sigma_{K-k}) \big)^K_{k=0}$ is calculated using the negated motion field $-\mathcal{M}$.
Although mean and covariance are calculated by forward propagating the transformation update from $0\rightarrow K$, the order of keyframes in $\mathcal{A}^{\text{bw}}$ is reversed, such that $\mathcal{A}^{\text{fw}}_0 = \mathcal{A}^{\text{bw}}_K$.
Due to $-\mathcal{M}$, $g$ moves backward within the motion field, which is reversed into forward flow direction.
Phenomenologically, this means motion sources from $\mathcal{A}^{\text{fw}}$ act as motion drains for $\mathcal{A}^{\text{bw}}$.
Note that a general inverse motion field $\mathcal{M}^{-1}$ is not available but $-\mathcal{M}$ offers a reasonable approximation and ensures the required symmetry~\cite{holynski2021animating}.
We visualize this in \cref{app:animation}.

\paragraph*{Opacity Blending.}
While $\mathcal{A}^{\text{fw}}$ and $\mathcal{A}^{\text{bw}}$ internally offer a smooth animation, looping both will result in a visible cut when Gaussians snap back to their initial state at the end of the sequence.
Furthermore, two instances of each dynamic Gaussian double the local scene density.
To mitigate these effects, we multiply opacity $o$ of each $g\in\mathcal{G}^{\text{dyn}}$ by $w_k^o = \sin^2\left(\frac{\pi k}{2K}\right)$.
Opacity for the forward and backward sequence is defined as $\mathcal{O}^{\text{fw}} = \big( ( 1 - w_k^o ) o \big)^K_{k=0}$ and $\mathcal{O}^{\text{bw}} = \big( w_k^o o \big)^K_{k=0}$.
As $\mathcal{O}^{\text{fw}}$ decreases from $o$ to $0$ and $\mathcal{O}^{\text{bw}}$ increases from $0$ to $o$, the backward instance fades in while the forward instance fades out.
This results in Gaussians being most visible in their original state while becoming less visible as they diverge due to applied transformations.
Note that $\mathcal{O}^{\text{fw}}_k + \mathcal{O}^{\text{bw}}_k = o$ for all $k$ ensures consistent local scene density and $\mathcal{O}^{\text{fw}}_K = \mathcal{O}^{\text{bw}}_0 = 0$ in combination with $\mathcal{A}^{\text{fw}}_0 = \mathcal{A}^{\text{bw}}_K$ ensures a seamless transition between looping sequences (cf.\ \cref{app:animation}).

\paragraph*{Randomized Animation States.}
Loops can still be perceivable due to large repeating patterns, which 2D cinemagraphy mitigates by warping deep features and inpainting.
For 3D Gaussians, we instead randomly offset the global animation state $t\in[0,1)$ of each $g$ by an individual offset $t^{\text{off}}\in[0,1)$.
The offset differs between Gaussians, but corresponding forward and backward instances are offset by the same $t^{\text{off}}$ to maintain the symmetric transformation of each individual $g$.
Note that the looping dynamic is not disrupted; only the starting point of each $g$ along its transformation trajectory is shifted.
It may be infeasible to completely disrupt patterns in specific fluid regions, so a maximum allowed random offset can be used to restrict $t^{\text{off}}$.

\subsubsection{Rendering}\label{sec:rendering}
We implement our CUDA renderer on top of an optimized 3DGS implementation~\cite{hahlbohm2026fastergs}.
Our symmetric animation technique requires rendering multiple instances of each $g\in\mathcal{G}^{\text{dyn}}$, each transformed by the keyframe of its own animation state.
For VRAM efficiency, these instances are virtual, i.e., share all animation-independent parameter buffers.
With the local animation state $\tau = (t + t^{\text{off}}) \bmod 1$, the keyframe of a virtual Gaussian instance is $k = \lfloor \tau K\rfloor$.
For smoother animation, means, covariances, and opacities of neighboring keyframes can be linearly interpolated with weight $\tau K - k$.
While linearly interpolating covariances is not exact with respect to rotation, the covariance difference between keyframes is typically small.
Consequently, it is a visually smooth and efficient alternative to proper rotation interpolation.

\section{Evaluation}
\label{sec:evaluation}
\begin{figure*}
\centering
\includegraphics[width=0.9\linewidth]{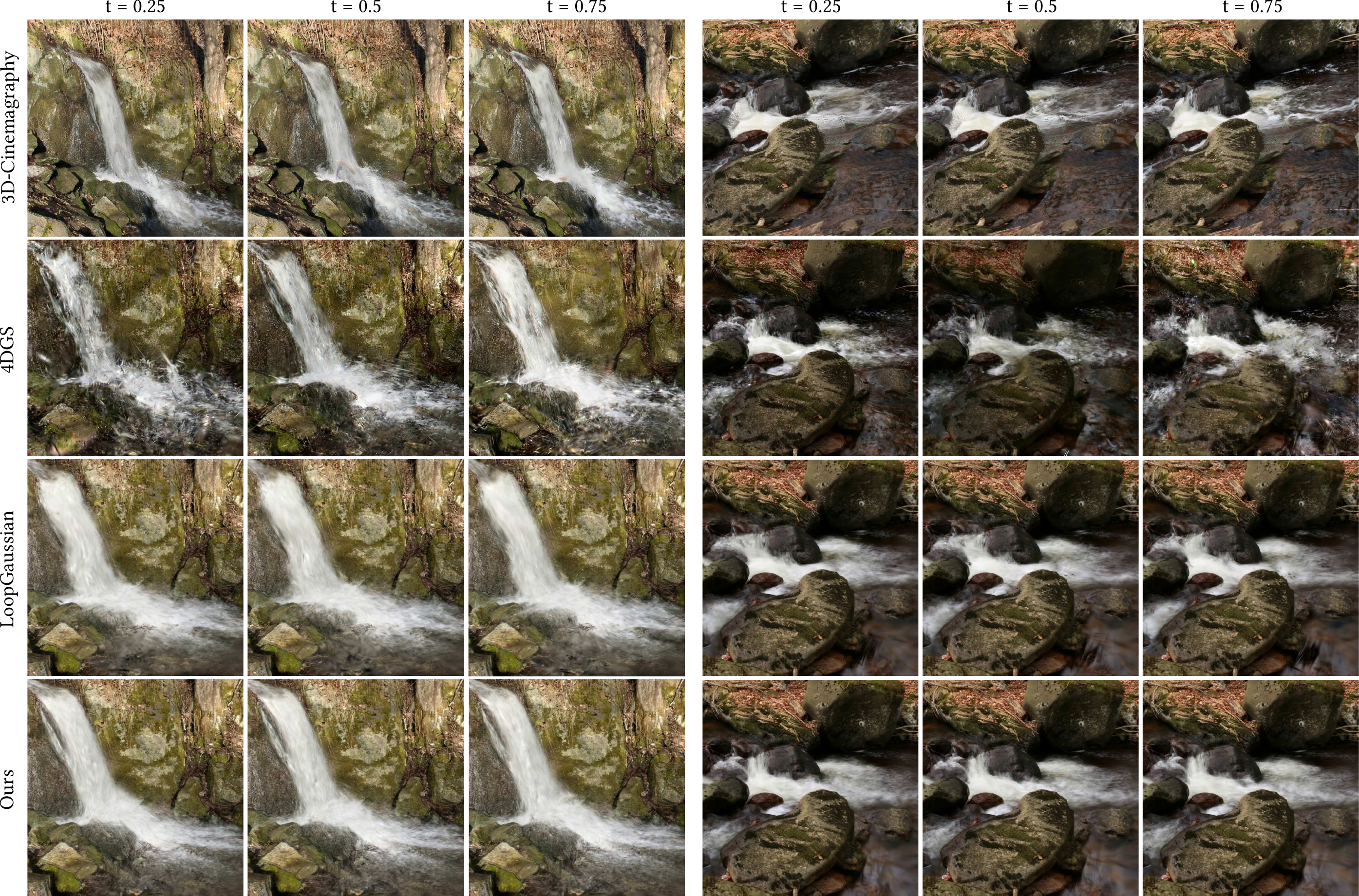}
\caption{%
Qualitative comparison with baselines for novel-view renderings.
Our method creates a visually plausible time-consistent fluid animation, while 3D-Cinemagraphy~\cite{li2023_3dcinemagraphy} introduces artifacts, 4DGS~\cite{yang20244dgaussians} fails to capture high-frequency water motion and generates a non-plausible fluid animation, and LoopGaussian~\cite{li2024loopgaussian} creates motion that does not follow realistic fluid flow (cf.\ our video).
}\label{fig:compare_rgb}
\end{figure*}

\begin{figure*}[t]
\centering
\includegraphics[width=1.0\linewidth]{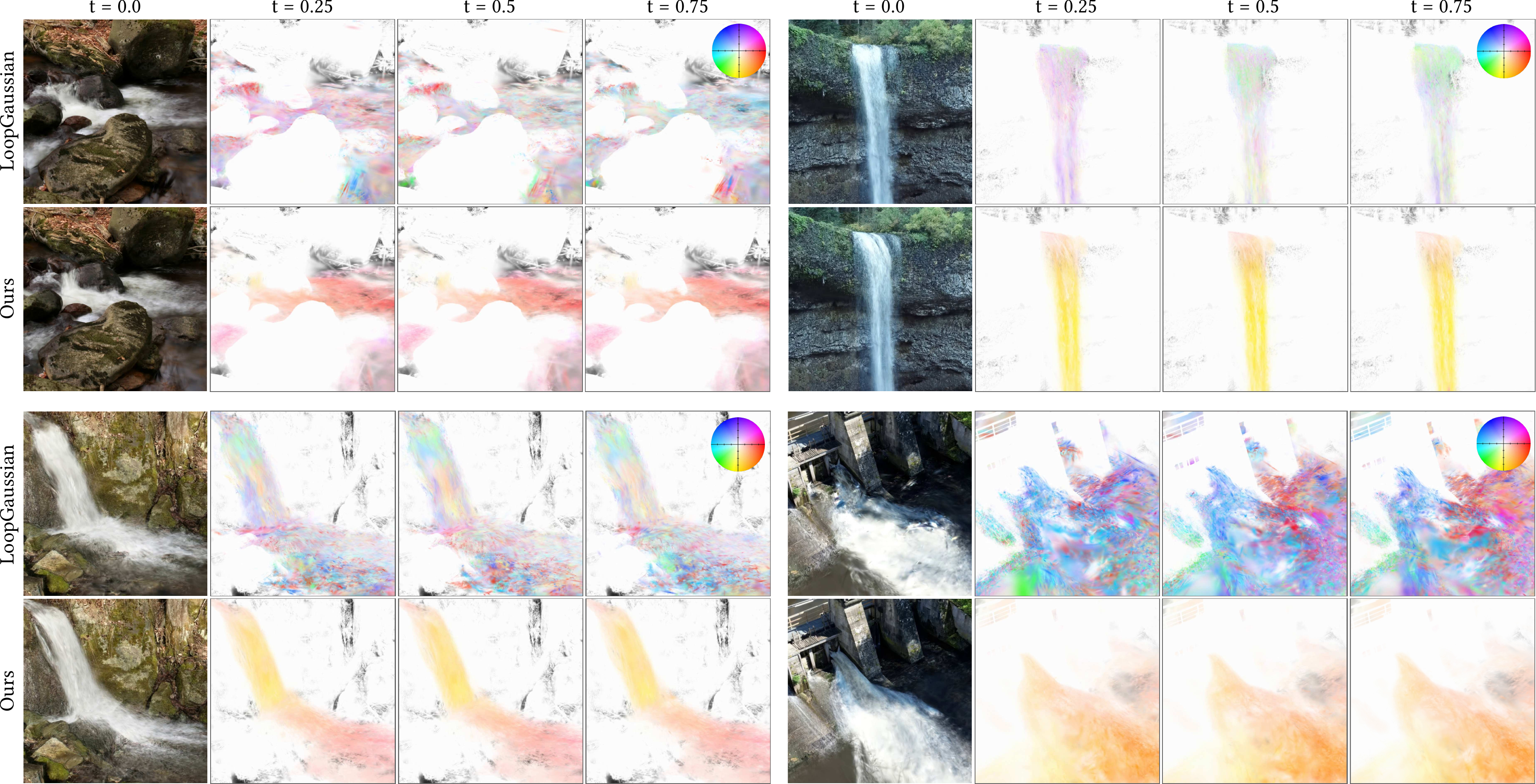}
\caption{%
Qualitative comparison between LoopGaussian~\cite{li2024loopgaussian} and our method on four scenes from our dataset where we visualize scene flow by coloring Gaussians according to their current motion~\cite{baker2007flowcolor}. We show that our method creates time-consistent motion that resembles the natural flow of water, while the motion synthesized by LoopGaussian points in arbitrary directions. The results were rendered from a novel viewpoint.
}\label{fig:compare_flow}
\end{figure*}

We evaluate our method and compare it to recent approaches with similar goals in a series of experiments.
We conduct all experiments on the same system using an RTX 4090 GPU.
Quantitatively evaluating synthesized fluid motion is inherently difficult: casually captured scenes provide no ground-truth motion, and static 3DGS represents moving water as a temporally averaged, blurred region whose high-frequency detail no method operating on a static reconstruction can recover without hallucination.
As our goal is visual plausibility rather than physical accuracy, and no suitable no-reference metrics for fluid motion exist, we assess motion quality qualitatively, primarily through our video, and evaluate efficiency quantitatively.

\subsection{Dataset}
As public benchmark datasets rarely feature fluid elements, we captured seven scenes containing waterfalls of different scales and shapes as well as a cascading creek and a weir.
All scenes were captured either by taking photographs with a DSLR camera or by recording multiple video clips with a drone.
We preprocess the raw data using \textit{COLMAP}~\cite{schoenberger2016colmap1} for camera parameter estimation, image undistortion, and scene alignment.
Next, we train a standard 3DGS model on 4\texttimes\ downsampled images using default hyperparameter settings~\cite{kerbl3Dgaussians, hahlbohm2026fastergs}.
In addition to this data, our dataset release will include several auxiliaries, such as pre-defined 3D motion hints, motion fields, and dynamic Gaussian segmentation masks (cf.\ \cref{tab:dataset}).
Per-scene motion field properties are provided in \cref{app:motion_field_prop}.
\begin{table}[tb]
\caption{%
Properties of all dataset scenes including number of total Gaussians $\left|\mathcal{G}\right|$, dynamic Gaussians $|\mathcal{G}^{\textit{dyn}}|$ and 3D motion hints $\left|\mathcal{H}\right|$.
}\label{tab:dataset}
\centering
\setlength\tabcolsep{8.1pt}
\scriptsize
\begin{tabular}{@{}lcccccc}
\toprule
Scene & \#Images & Resolution & Camera & $\left|\mathcal{G}\right|$  & $|\mathcal{G}^{\textit{dyn}}|$ & $\left|\mathcal{H}\right|$ \\
\midrule
Creek & 103 & 6223\texttimes4080 & DSLR & 1.59M & 51K & 11 \\
Frogfall & 186 & 6223\texttimes4080 & DSLR & 6.72M & 26K & 12 \\
Silverfalls & 67 & 3811\texttimes2144 & Drone & 5.62M & 51K & 10 \\
Smallfall & 74 & 6223\texttimes4080 & DSLR & 5.56M & 87K & 8 \\
Town & 159 & 6223\texttimes4080 & DSLR & 3.11M & 12K & 8 \\
Waterfall & 158 & 6223\texttimes4080 & DSLR & 5.26M & 182K & 14 \\
Weir & 333 & 2624\texttimes1475 & Drone & 5.16M & 162K & 14 \\
\bottomrule
\end{tabular}
\end{table}

\subsection{Baselines}
\begin{figure*}[t]
\centering
\includegraphics[width=1.0\linewidth]{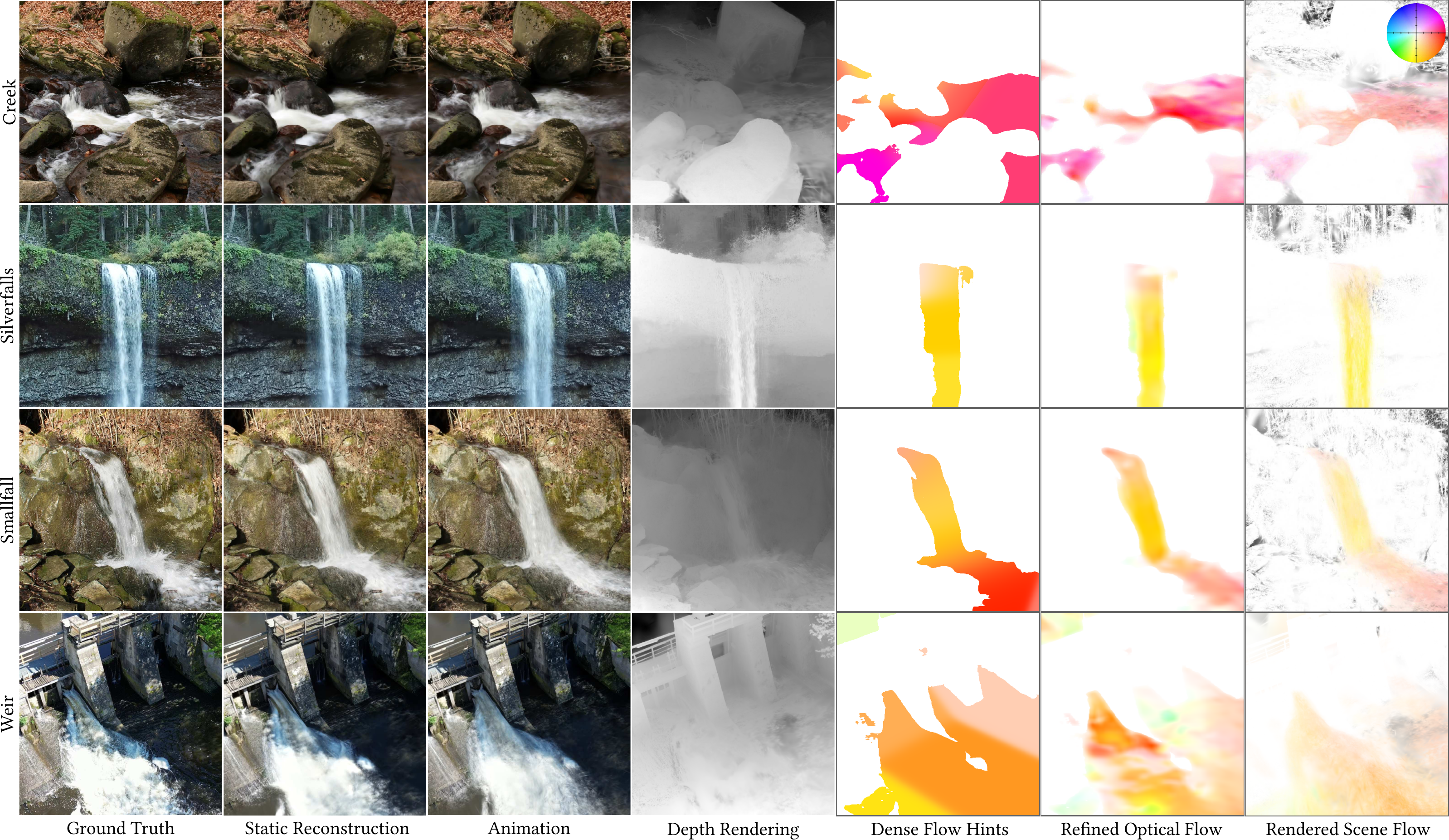}
\caption{%
Ground truth images along with renderings of the static and animated 3DGS reconstruction, depth renderings, dense flow hints $\mathcal{F}^{\text{dense}}_n$, refined optical flow $\mathcal{F}_n$, and scene flow by coloring Gaussians according to their current motion~\cite{baker2007flowcolor}.
Depicted camera poses are used as closest input poses for novel-view renderings (cf. \cref{fig:compare_rgb} and \cref{fig:compare_flow}).
}\label{fig:method_stages}
\end{figure*}

We compare our method with \textit{4DGS}~\cite{yang20244dgaussians}, \textit{LoopGaussian}~\cite{li2024loopgaussian}, and \textit{3D-Cinemagraphy}~\cite{li2023_3dcinemagraphy}.
We train \textit{4DGS} models at the same resolution as our 3DGS models and limit the maximum number of Gaussians to $6.5$M to prevent VRAM issues.
For fair comparison and precise evaluation, we integrate the \textit{LoopGaussian} method, i.e., their trained 3D motion field MLP and their symmetric animation technique into our framework and GUI.
\textit{LoopGaussian} also relies on Gaussian segmentation to determine dynamic scene regions, which we generate based on our motion field similar to how we separate static and dynamic Gaussians in our method.
For \textit{3D-Cinemagraphy}, we render novel views by performing their layered depth image (LDI) reconstruction on the downsampled RGB image of the closest dataset view.
As an additional required input, we project our 3D motion hints based on the input pose to obtain sparse flow hints.
We note that achieving identical camera poses within our 3DGS model and the LDI of \textit{3D-Cinemagraphy} is challenging since depth information is acquired by monocular depth estimation, resulting in significantly different scene scales.
This effect grows with the distance between novel and input pose and thus complicates free-viewpoint exploration, our target application.

For all comparisons, we render a single $4$-second animation loop of $240$ frames in total.
We limit \textit{3D-Cinemagraphy} to $60$ frames, since we find that using more tends to degrade the quality of their animation.
For each scene, we render from a novel viewpoint diverging from the closest input pose both in camera translation and rotation.

\subsection{Qualitative Comparisons}
\begin{figure}[tb]
\centering
\includegraphics[width=1.0\linewidth]{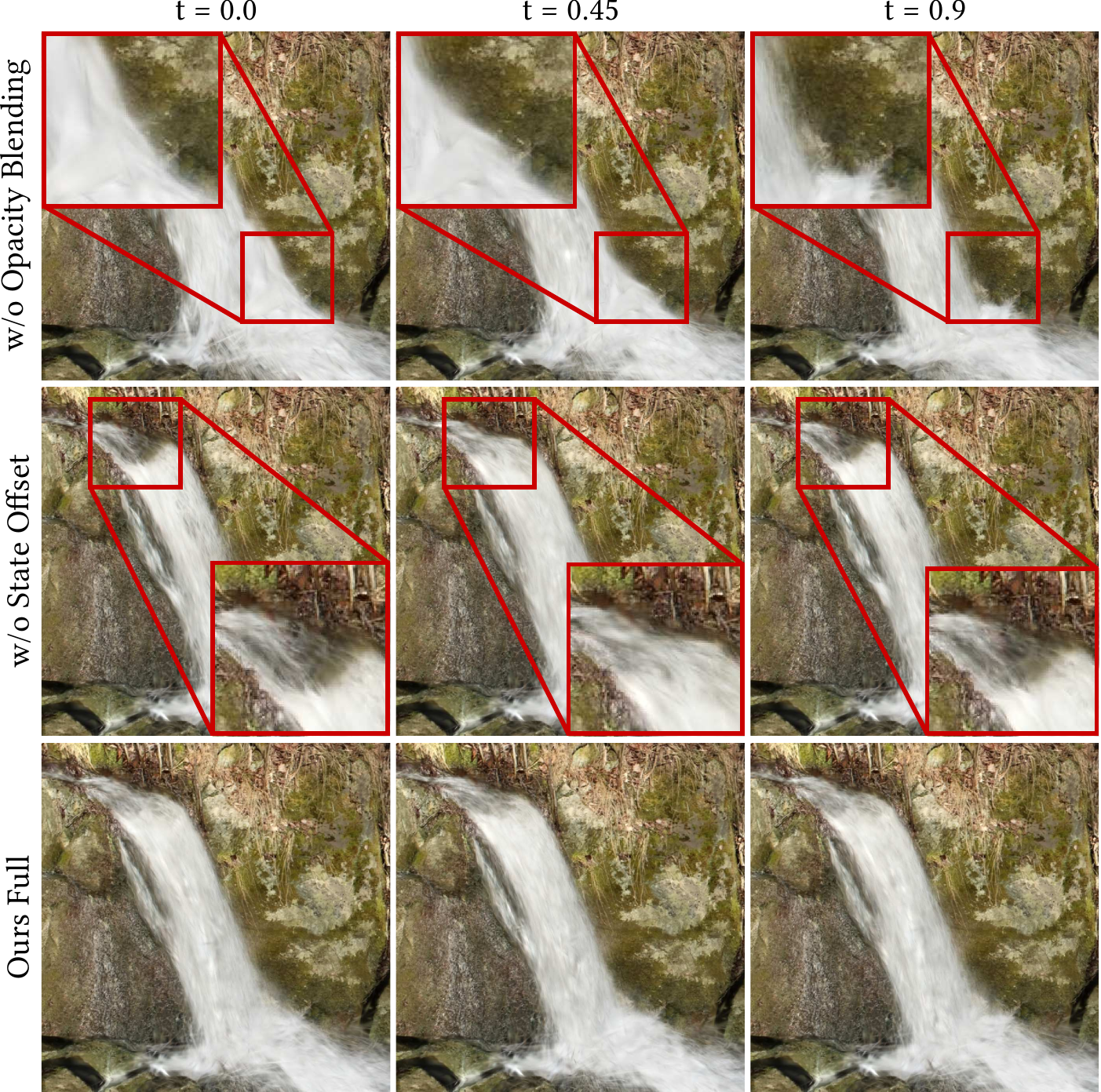}
\caption{%
Ablation of opacity blending and state offset. We show that both techniques are required for creating a seamless loop without noticeable cuts (top) and reoccurring patterns (middle).
}\label{fig:ablation_opacity_offset}
\end{figure}

\begin{figure}[tb]
\centering
\includegraphics[width=1.0\linewidth]{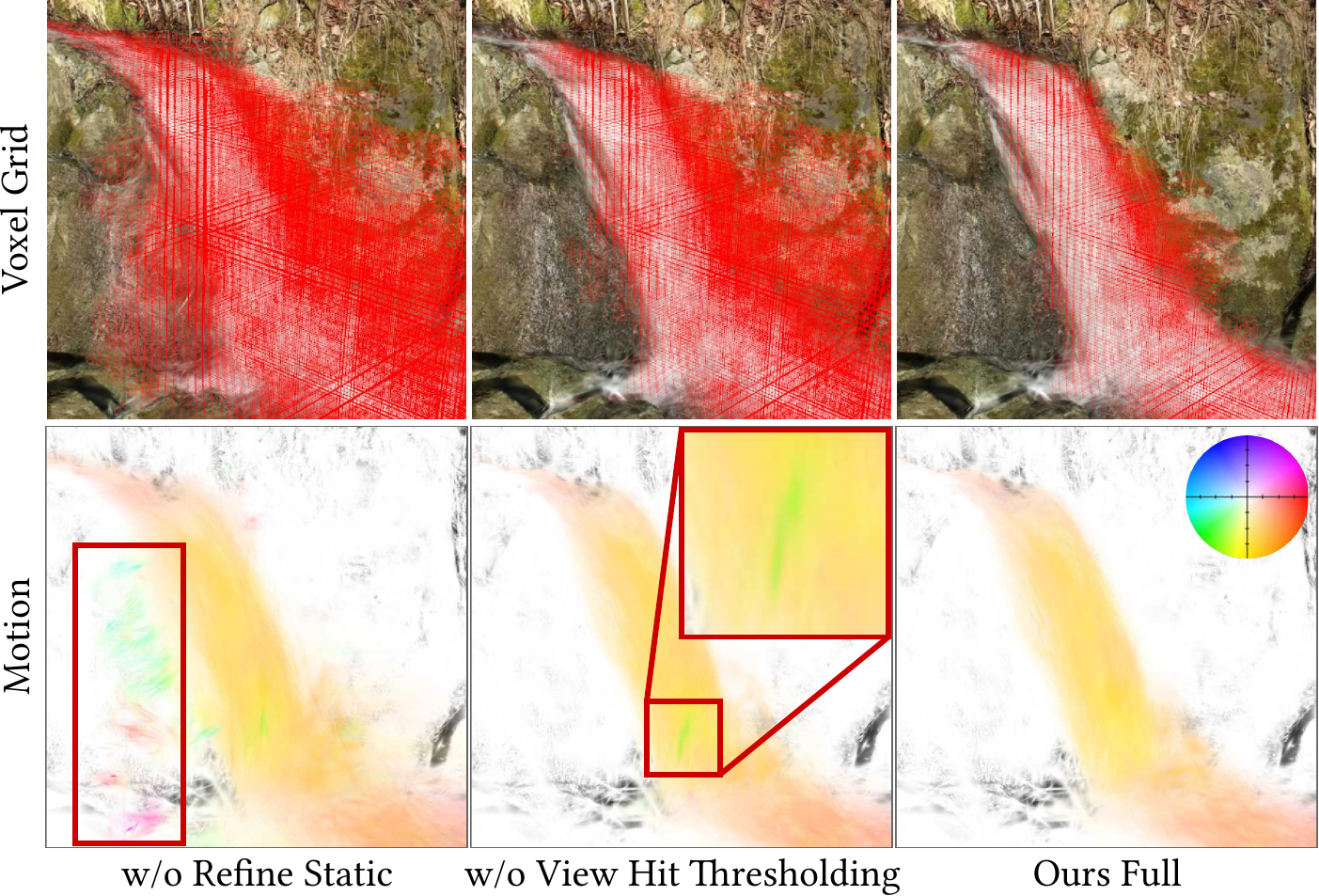}
\caption{%
Ablation of motion field refinements. The top row shows a voxel grid visualization, and the bottom row Gaussians colored based on their current motion. By refining static regions and view hit thresholding, we can eliminate motion in static regions (left) and wrong motion (middle).
}\label{fig:ablation_motion_field}
\end{figure}

\begin{figure}[tb]
\centering
\includegraphics[width=1.0\linewidth]{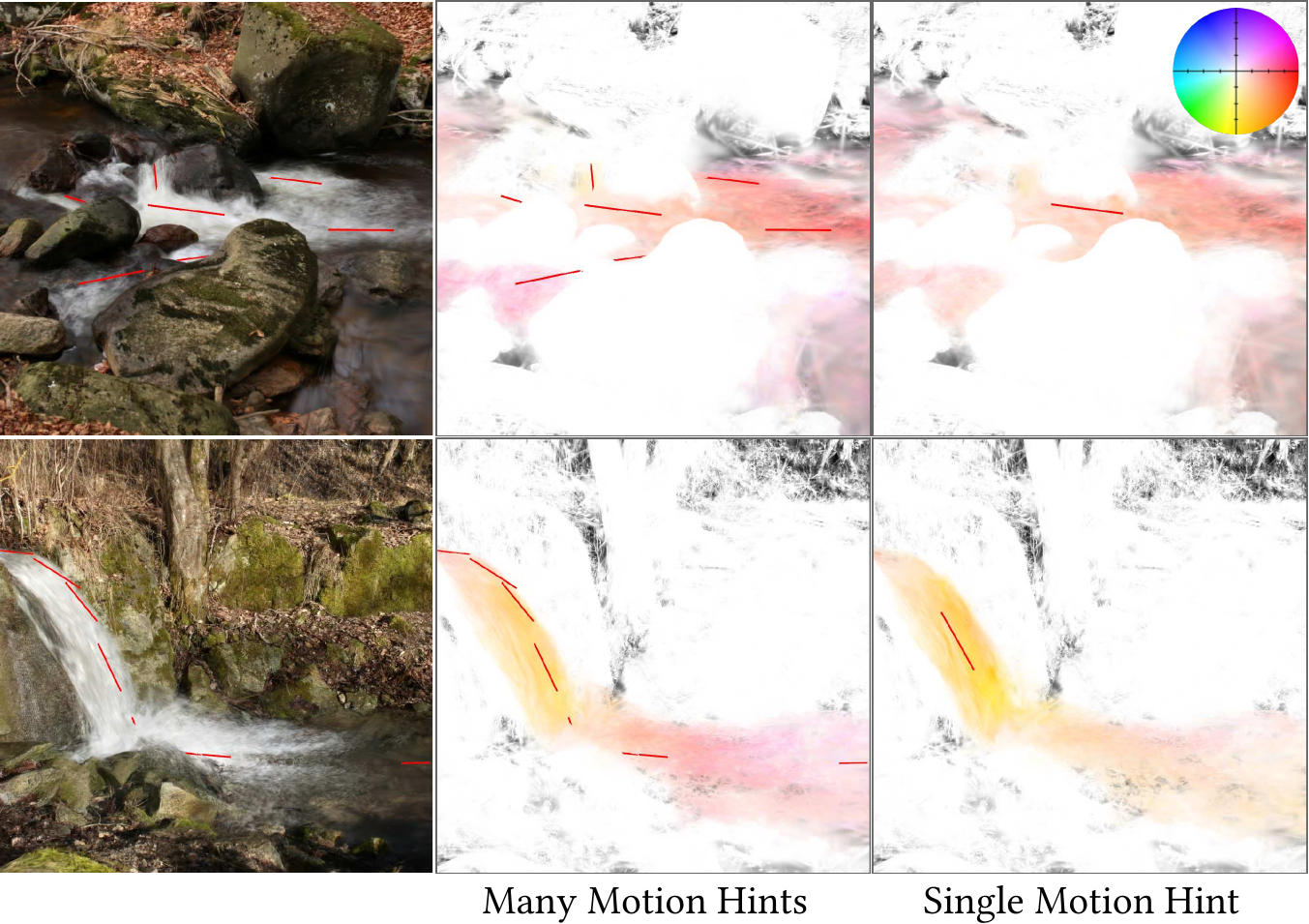}
\caption{%
We show that on some scenes (top) a single directional 3D motion hint can produce nearly the same motion as many hints. On other scenes (bottom), one hint does not suffice to model the transition from a falling waterfall into a flowing river. Hints are illustrated as red lines.
}\label{fig:ablation_hints}
\end{figure}

\cref{fig:compare_rgb} presents qualitative comparisons of rendered animation frames for all methods.
3D-Cinemagraphy produces visually compelling animations for viewpoints close to the input image, as it largely relies on high-detail 2D cinemagraph synthesis.
However, as the viewpoint deviates further from the reference pose, limitations of the underlying layered depth image representation become increasingly apparent, leading to geometric inconsistencies, animation artifacts, and missing content not observed in the reference view.
In contrast, Gaussian splatting-based approaches maintain viewpoint-consistent scene geometry, making visual quality primarily dependent on the reconstruction and animation of fluid regions.
Under casual capture conditions, 4DGS struggles to reconstruct high-frequency water dynamics, resulting in unstable and visually implausible motion.
As 4DGS models the entire scene as dynamic, it also fails to converge to the static solution in some regions (e.g., the darker rocks in \cref{fig:compare_rgb}).
LoopGaussian preserves a temporally stable appearance, but its motion does not follow coherent fluid flow and instead exhibits locally inconsistent warping behavior (best seen in our video).
To further analyze this effect, \cref{fig:compare_flow} visualizes the time-dependent motion of individual Gaussians by encoding projected motion direction as color.
LoopGaussian produces motion that varies both spatially and temporally within similar fluid regions, leading to incoherent flow patterns.
This behavior stems from its symmetric interpolation scheme, which blends forward- and backward-transformed Gaussian positions instead of rendering them separately.
In contrast, our method generates temporally consistent directional motion that more closely resembles plausible fluid flow.
\cref{fig:method_stages} shows the corresponding closest input views together with intermediate results of our pipeline: sparse hints are refined into detailed flow maps that follow the water, and the fused motion field reproduces this flow when rendered.
Our video further compares our animation side by side with the drone-captured footage used for reconstructing the 3DGS model of the ``Weir'' scene.

\begin{table}[tb]
\caption{%
Quantitative comparison for rendering one animation loop.
}\label{tab:comp_online}
\centering
\setlength\tabcolsep{3.3pt}
\scriptsize
\begin{tabular}{@{}lcccccccc|cc}
\toprule
  & \multicolumn{2}{c}{3D-Cinema.} & \multicolumn{2}{c}{4DGS} & \multicolumn{2}{c}{LoopGaussian} & \multicolumn{2}{c|}{Ours} & \multicolumn{2}{c}{3DGS} \\
Scene & VRAM & FPS & VRAM & FPS & VRAM & FPS & VRAM & FPS & VRAM & FPS \\
\midrule
Creek & 10.5\,GiB & <1 & 2.6\,GiB & 431 & 2.1\,GiB & 429 & 1.3\,GiB & 457 & 0.9\,GiB & 650 \\
Frogfall & 10.8\,GiB & <1 & 2.8\,GiB & 350 & 8.7\,GiB & 275 & 5.5\,GiB & 282 & 2.9\,GiB & 283 \\
Silverfalls & 13.6\,GiB & <1 & 2.7\,GiB & 301 & 7.3\,GiB & 282 & 4.6\,GiB & 283 & 2.4\,GiB & 328 \\
Smallfall & 10.5\,GiB & <1 & 2.7\,GiB & 430 & 7.2\,GiB & 349 & 4.5\,GiB & 342 & 2.3\,GiB & 389 \\
Town & 10.4\,GiB & <1 & 2.6\,GiB & 432 & 4.0\,GiB & 409 & 2.5\,GiB & 407 & 1.4\,GiB & 569 \\
Waterfall & 10.4\,GiB & <1 & 2.7\,GiB & 357 & 6.8\,GiB & 186 & 4.3\,GiB & 245 & 2.4\,GiB & 296 \\
Weir & \;\,6.1\,GiB & <1 & 2.6\,GiB & 440 & 6.7\,GiB & 326 & 4.2\,GiB & 312 & 1.8\,GiB & 393 \\
\midrule
Average & 10.3\,GiB & <1 & 2.7\,GiB & 392 & 6.1\,GiB & 322 & 3.8\,GiB & 333 & 2.0\,GiB & 416 \\
\bottomrule
\end{tabular}
\end{table}

\subsection{Quantitative Comparisons}
In \cref{tab:comp_online}, we compare rendering performance and VRAM consumption across methods.
All Gaussian Splatting-based methods render interactively at over $300$\,FPS on average.
Between these, 4DGS exhibits the lowest average memory consumption at $2.7$\,GiB, followed by our method at $3.8$\,GiB.
LoopGaussian requires considerably more memory in several scenes, reaching $8.7$\,GiB for ``Frogfall'' compared to $5.5$\,GiB for our approach.
This overhead is primarily caused by the per-Gaussian MLP evaluations required for motion prediction.
3D-Cinemagraphy is neither real-time capable nor memory efficient.
We also show static 3DGS for reference.

\begin{table}[tb]
\caption{%
Quantitative comparison of motion field preprocessing. For our method, we also show storage cost and sub-task timings.
}\label{tab:comp_preprocess}
\centering
\setlength\tabcolsep{7pt}
\scriptsize
\begin{tabular}{@{}lc|ccccc}
\toprule
 & LoopGaussian & \multicolumn{5}{c}{Ours} \\
Scene & Total Time & Total Time & Masks & Flow & Fusion & Storage \\
\midrule
Creek & 3m42s & 40s & 22s & 16s & 2s & $5.2$\,MB \\
Frogfall & 1m55s & 1m03s & 38s & 17s & 8s & $44$\,KB \\
Silverfalls & 3m26s & 37s & 22s & 13s & 2s & $0.2$\,MB \\
Smallfall & 6m25s & 28s & 17s & 9s & 2s & $0.7$\,MB \\
Town & 1m01s & 55s & 33s & 16s & 6s & $1.1$\,MB \\
Waterfall & 13m20s & 57s & 32s & 18s & 7s & $4.0$\,MB \\
Weir & 12m15s & 1m42s & 1m01s & 35s & 5s & $3.4$\,MB \\
\midrule
Average & 6m01s & 55s & 32s & 18s & 5s & $2.1$\,MB \\
\bottomrule
\end{tabular}
\end{table}

\cref{tab:comp_preprocess} further evaluates the one-time cost for constructing the 3D motion field.
Our method requires approximately one minute on average, with runtime scaling primarily with the number of processed input views.
The dominant cost stems from image segmentation.
LoopGaussian exhibits substantially higher preprocessing times, averaging around six minutes and reaching roughly thirteen minutes for the ``Waterfall'' scene.
Notably, we provide the required static/dynamic Gaussian segmentation for LoopGaussian directly from our motion field pipeline; obtaining such a segmentation independently, e.g., using SAGA~\cite{cen2025saga}, would introduce additional preprocessing overhead.
The peak memory usage of our fusion stage is $13.4$\,GiB, which can easily be bounded by limiting the initial voxel grid resolution.
Despite relying on a dense voxel grid during fusion, the resulting sparse motion field requires only $2$\,MB on average.

\subsection{Ablation Study}
In \cref{fig:ablation_opacity_offset} (top row), we render our animation without opacity blending and find that start and end frames of one animation sequence differ visually.
Looping such sequences results in a visible cut due to the backward instance snapping back to its starting position.
We also show the benefit of our proposed random state offset in \cref{fig:ablation_opacity_offset} (middle row), where the top of the waterfall creates a recurring pattern in the sequence when such state offset is disabled.
In \cref{fig:ablation_motion_field}, we analyze our motion field refinement techniques.
The left column clearly shows the effectiveness of motion reduction within static scene regions to prevent visual artifacts of unwanted movement caused by incorrectly reprojected flow pixels.
The middle column shows that our view hit thresholding reduces incoherent motion of dynamic scene regions due to an ill-posed reprojection problem (bottom image).
It also further reduces voxels within empty scene regions to better fit the shape of the dynamic Gaussian region (top-middle compared to the top-right image).
We also evaluate the use of many 3D motion hints compared to a single directional hint in \cref{fig:ablation_hints}.
In the top row, we show that for some scenes a single motion hint offers enough guidance for the flow estimation network to produce motion of similar quality to using multiple hints.
Other scenes require additional hints to capture the transition from water falling down the waterfall to flowing along the river.
In practice, hints can be added where the animation deviates from the expected flow, as demonstrated for ``Smallfall'' in our video.

\section{Discussion, Limitations, and Future Work}
\label{sec:discussion}

\begin{figure}[tb]
\centering
\includegraphics[width=1.0\linewidth]{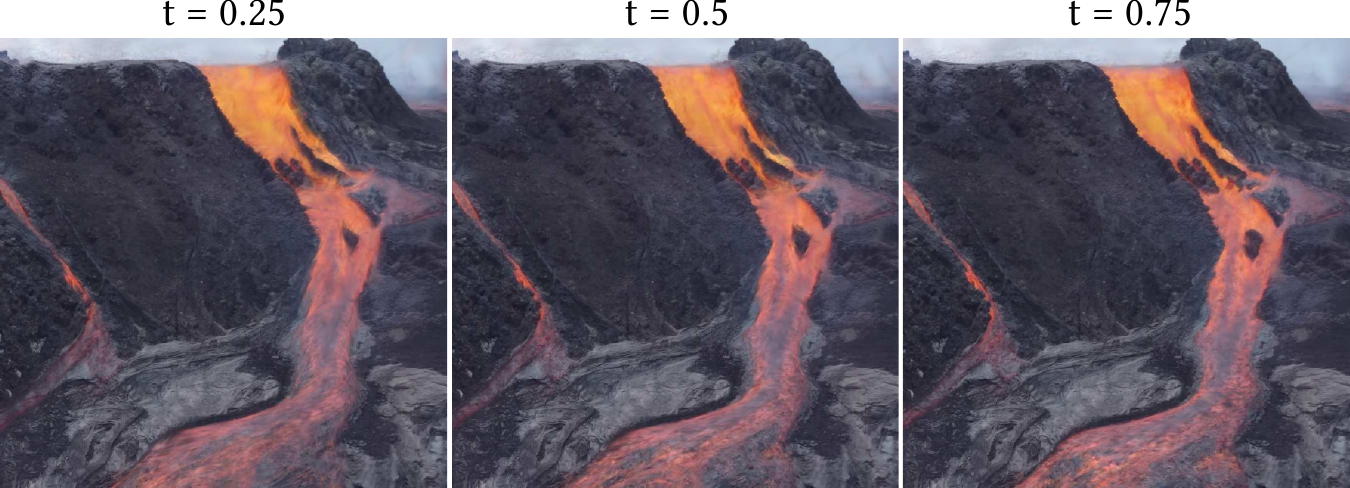}
\caption{%
Our method generalizes to fluids other than water, such as lava. We show the animation from a novel camera trajectory in our video.
}\label{fig:volcano}
\end{figure}

Our method demonstrates that visually plausible fluid animation can be efficiently integrated into static 3DGS scenes without requiring a fully dynamic reconstruction.
While we primarily focus on water, the proposed approach also generalizes to other fluids such as lava (\cref{fig:volcano}).

Since the 3D motion field is precomputed only once and stored compactly, it could be distributed alongside static 3DGS scenes and rendered efficiently even on resource-constrained devices.
Furthermore, the optical flow estimation process is flexible with respect to the input views.
The RGB inputs $\mathcal{I}_{1,\cdots,N}$ may consist of original training images, renderings from user-defined viewpoints, or a combination of both, enabling targeted refinement of motion in challenging scene regions through our interactive GUI.

\begin{figure}[tb]
\centering
\includegraphics[width=1.0\linewidth]{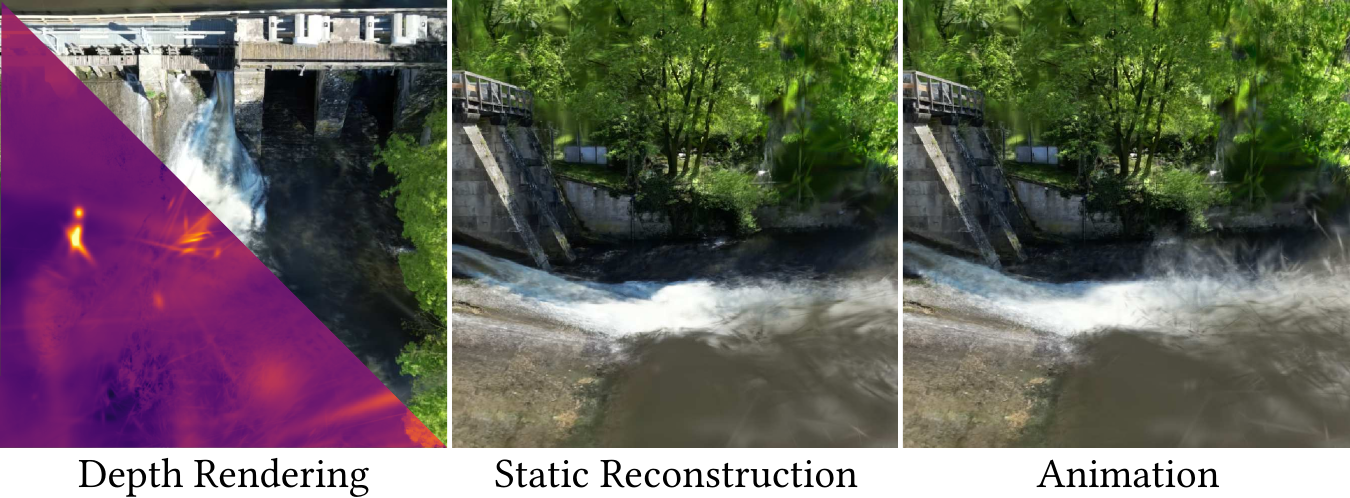}
\caption{%
Failure case on ``Weir'': inaccurate Gaussian geometry (depth, left) and a poorly reconstructed water surface (middle) lead to wrongly reprojected flow and thus an inaccurate motion field, visible as animation artifacts (right).
}\label{fig:limitations}
\end{figure}

Despite these advantages, our approach is not without limitations.
First, the quality of the generated motion field strongly depends on the underlying 3DGS reconstruction.
Depth inconsistencies or reconstruction artifacts can lead to inaccurate motion fusion and visible animation artifacts (\cref{fig:limitations}).

Similarly, the method assumes that the fluid surface itself is reconstructed reasonably well.
This assumption may fail for highly reflective or visually ambiguous water surfaces, where the 3DGS reconstruction does not create enough Gaussians.
Our method also cannot recover fluid detail that is absent from the reconstructed 3DGS representation and, as depth stems from Gaussians at the visible water surface, motion is largely confined to this surface.

Future work could explore dynamic updates of the motion field to enable interactive or physically responsive fluid behavior.
This could allow animated fluids to react to user interaction or external scene events while maintaining real-time performance.

\section{Conclusion}
We presented a method for animating water in static 3D Gaussian Splatting scenes using cinemagraph-inspired motion synthesis.
Our approach lifts 2D fluid motion into a coherent 3D Eulerian motion field and applies it to Gaussian primitives to generate seamless looping animations that preserve real-time rendering and free-viewpoint navigation.
In contrast to dynamic reconstruction approaches, our method operates on casually captured static scenes and does not require dynamic multi-view acquisition or costly physical simulation.
Experiments on a newly captured dataset show more plausible water motion than prior methods at low computational cost.
We believe that combining static scene reconstruction with efficient motion synthesis opens promising directions for immersive, controllable, and realistic 3D content creation.

\FloatBarrier

\section*{Acknowledgments}
We would like to thank Brian Scotland and Junbo Li for their valuable contributions to our dataset, as well as Philipp Gronek for his assistance with the comparisons.
This work was partially funded by the DFG projects ``Real-Action VR'' (ID 523421583) and ``Increasing Realism of Omnidirectional Videos in Virtual Reality'' (ID 491805996).
We also acknowledge funding from the Lower Saxony Ministry of Science and Culture under grant number ZN3994 within the Lower Saxony ``Vorab'' of the Volkswagen Foundation and support by the Center for Digital Innovations (ZDIN).

\bibliographystyle{eg-alpha-doi}
\bibliography{references}

\newcommand{\etalchar}[1]{$^{#1}$}
\begin{thebibliography}{\uppercase{PCPMMN21}}

\bibitem[BRS{\etalchar{*}}07]{baker2007flowcolor}
\textsc{Baker S., Roth S., Scharstein D., Black M.~J., Lewis J., Szeliski R.}:
\newblock A database and evaluation methodology for optical flow.
\newblock In \emph{ICCV} (2007), pp.~1--8.
\newblock \href {https://doi.org/10.1109/ICCV.2007.4408903}
  {\path{doi:10.1109/ICCV.2007.4408903}}.

\bibitem[CFY{\etalchar{*}}25]{cen2025saga}
\textsc{Cen J., Fang J., Yang C., Xie L., Zhang X., Shen W., Tian Q.}:
\newblock Segment any {3D Gaussians}.
\newblock In \emph{AAAI} (USA, 2025), AAAI Press, pp.~1971--1979.
\newblock \href {https://doi.org/10.1609/aaai.v39i2.32193}
  {\path{doi:10.1609/aaai.v39i2.32193}}.

\bibitem[CGH{\etalchar{*}}26]{carion2026sam3}
\textsc{Carion N., Gustafson L., Hu Y.-T., Debnath S., Hu R., Coll-Vinent
  D.~S., Ryali C., Alwala K.~V., Khedr H., et~al. A.~H.}:
\newblock {SAM} 3: Segment anything with concepts.
\newblock In \emph{ICLR} (2026).
\newblock URL: \url{https://openreview.net/forum?id=r35clVtGzw}.

\bibitem[CJ23]{cao2023hexplane}
\textsc{Cao A., Johnson J.}:
\newblock {HexPlane}: A fast representation for dynamic scenes.
\newblock In \emph{CVPR} (2023), pp.~130--141.
\newblock \href {https://doi.org/10.1109/CVPR52729.2023.00021}
  {\path{doi:10.1109/CVPR52729.2023.00021}}.

\bibitem[COP{\etalchar{*}}24]{cheng2024cutie}
\textsc{Cheng H.~K., Oh S.~W., Price B., Lee J.-Y., Schwing A.}:
\newblock Putting the object back into video object segmentation.
\newblock In \emph{CVPR} (2024), pp.~3151--3161.
\newblock \href {https://doi.org/10.1109/CVPR52733.2024.00304}
  {\path{doi:10.1109/CVPR52733.2024.00304}}.

\bibitem[CXG{\etalchar{*}}22]{Chen2022tensorf}
\textsc{Chen A., Xu Z., Geiger A., Yu J., Su H.}:
\newblock {TensoRF}: Tensorial radiance fields.
\newblock In \emph{ECCV} (2022), pp.~333--350.
\newblock \href {https://doi.org/10.1007/978-3-031-19824-3_20}
  {\path{doi:10.1007/978-3-031-19824-3_20}}.

\bibitem[DTM96]{debevec1996mar}
\textsc{Debevec P.~E., Taylor C.~J., Malik J.}:
\newblock Modeling and rendering architecture from photographs: a hybrid
  geometry- and image-based approach.
\newblock In \emph{SIGGRAPH} (1996), p.~11–20.
\newblock \href {https://doi.org/10.1145/237170.237191}
  {\path{doi:10.1145/237170.237191}}.

\bibitem[EKK19]{endo2019animatinglandscape}
\textsc{Endo Y., Kanamori Y., Kuriyama S.}:
\newblock Animating landscape: self-supervised learning of decoupled motion and
  appearance for single-image video synthesis.
\newblock \emph{ACM TOG 38}, 6 (2019).
\newblock \href {https://doi.org/10.1145/3355089.3356523}
  {\path{doi:10.1145/3355089.3356523}}.

\bibitem[FAEB26]{fiebelman2026letsnow}
\textsc{Fiebelman G., Averbuch-Elor H., Benaim S.}:
\newblock Let it snow! animating {3D Gaussian} scenes with dynamic weather
  effects via physics-guided score distillation.
\newblock In \emph{CVPR} (2026), pp.~37322--37331.

\bibitem[FFS{\etalchar{*}}25]{feng2025splashing}
\textsc{Feng Y., Feng X., Shang Y., Jiang Y., Yu C., Zong Z., Shao T., Wu H.,
  Zhou K., Jiang C., Yang Y.}:
\newblock {Gaussian Splashing}: Unified particles for versatile motion
  synthesis and rendering.
\newblock In \emph{CVPR} (2025), pp.~518--529.
\newblock \href {https://doi.org/10.1109/CVPR52734.2025.00057}
  {\path{doi:10.1109/CVPR52734.2025.00057}}.

\bibitem[FKMW{\etalchar{*}}23]{fridovichkeil2023kplanes}
\textsc{Fridovich-Keil S., Meanti G., Warburg F.~R., Recht B., Kanazawa A.}:
\newblock {K-Planes}: Explicit radiance fields in space, time, and appearance.
\newblock In \emph{CVPR} (2023), pp.~12479--12488.
\newblock \href {https://doi.org/10.1109/CVPR52729.2023.01201}
  {\path{doi:10.1109/CVPR52729.2023.01201}}.

\bibitem[FKYT{\etalchar{*}}22]{yu2021plenoxels}
\textsc{Fridovich-Keil S., Yu A., Tancik M., Chen Q., Recht B., Kanazawa A.}:
\newblock Plenoxels: Radiance fields without neural networks.
\newblock In \emph{CVPR} (2022), pp.~5491--5500.
\newblock \href {https://doi.org/10.1109/CVPR52688.2022.00542}
  {\path{doi:10.1109/CVPR52688.2022.00542}}.

\bibitem[FYW{\etalchar{*}}22]{fang2022tineuvox}
\textsc{Fang J., Yi T., Wang X., Xie L., Zhang X., Liu W., Nie\ss{}ner M., Tian
  Q.}:
\newblock Fast dynamic radiance fields with time-aware neural voxels.
\newblock In \emph{SIGGRAPH Asia} (2022).
\newblock \href {https://doi.org/10.1145/3550469.3555383}
  {\path{doi:10.1145/3550469.3555383}}.

\bibitem[GGSC96]{gortler1996lumigraph}
\textsc{Gortler S.~J., Grzeszczuk R., Szeliski R., Cohen M.~F.}:
\newblock The lumigraph.
\newblock In \emph{SIGGRAPH} (1996), pp.~43--54.
\newblock \href {https://doi.org/10.1145/237170.237200}
  {\path{doi:10.1145/237170.237200}}.

\bibitem[GXG18]{gao2018im2flow}
\textsc{Gao R., Xiong B., Grauman K.}:
\newblock {Im2Flow}: Motion hallucination from static images for action
  recognition.
\newblock In \emph{CVPR} (2018), pp.~5937--5947.
\newblock \href {https://doi.org/10.1109/CVPR.2018.00622}
  {\path{doi:10.1109/CVPR.2018.00622}}.

\bibitem[HCSS21]{holynski2021animating}
\textsc{Holynski A., Curless B., Seitz S.~M., Szeliski R.}:
\newblock Animating pictures with {Eulerian} motion fields.
\newblock In \emph{CVPR} (2021), pp.~5806--5815.
\newblock \href {https://doi.org/10.1109/CVPR46437.2021.00575}
  {\path{doi:10.1109/CVPR46437.2021.00575}}.

\bibitem[HFEM26]{hahlbohm2026fastergs}
\textsc{Hahlbohm F., Franke L., Eisemann M., Magnor M.}:
\newblock {Faster-GS}: Analyzing and improving {Gaussian} splatting
  optimization.
\newblock In \emph{CVPR} (2026), pp.~18946--18957.

\bibitem[HHB18]{hao2018vidgen}
\textsc{Hao Z., Huang X., Belongie S.}:
\newblock Controllable video generation with sparse trajectories.
\newblock In \emph{CVPR} (2018), pp.~7854--7863.
\newblock \href {https://doi.org/10.1109/CVPR.2018.00819}
  {\path{doi:10.1109/CVPR.2018.00819}}.

\bibitem[HHV{\etalchar{*}}21]{halperin2021loops}
\textsc{Halperin T., Hakim H., Vantzos O., Hochman G., Benaim N., Sassy L.,
  Kupchik M., Bibi O., Fried O.}:
\newblock Endless loops: detecting and animating periodic patterns in still
  images.
\newblock \emph{ACM TOG 40}, 4 (2021).
\newblock \href {https://doi.org/10.1145/3450626.3459935}
  {\path{doi:10.1145/3450626.3459935}}.

\bibitem[HZ03]{hartley2003mvcv}
\textsc{Hartley R., Zisserman A.}:
\newblock \emph{Multiple View Geometry in Computer Vision}, 2~ed.
\newblock Cambridge University Press, New York, NY, 2003.
\newblock URL: \url{https://dl.acm.org/doi/10.5555/861369}.

\bibitem[JMG{\etalchar{*}}25]{jia2025physic}
\textsc{Jia E. Y.-T., Mao J., Gao Z., Zhao Y., Wang Y.}:
\newblock Learning an implicit physics model for image-based fluid simulation.
\newblock In \emph{ICCV} (2025), pp.~7055--7057.
\newblock \href {https://doi.org/10.1109/ICCV51701.2025.00662}
  {\path{doi:10.1109/ICCV51701.2025.00662}}.

\bibitem[JU97]{julier1997ut}
\textsc{Julier S.~J., Uhlmann J.~K.}:
\newblock New extension of the {Kalman} filter to nonlinear systems.
\newblock In \emph{Signal Processing, Sensor Fusion, and Target Recognition VI}
  (USA, 1997), vol.~3068, SPIE, pp.~182--193.
\newblock \href {https://doi.org/10.1117/12.280797}
  {\path{doi:10.1117/12.280797}}.

\bibitem[KKLD23]{kerbl3Dgaussians}
\textsc{Kerbl B., Kopanas G., Leimkuehler T., Drettakis G.}:
\newblock {3D Gaussian} splatting for real-time radiance field rendering.
\newblock \emph{ACM TOG 42}, 4 (2023).
\newblock \href {https://doi.org/10.1145/3592433} {\path{doi:10.1145/3592433}}.

\bibitem[LCS{\etalchar{*}}23]{li2023_3dcinemagraphy}
\textsc{Li X., Cao Z., Sun H., Zhang J., Xian K., Lin G.}:
\newblock {3D} cinemagraphy from a single image.
\newblock In \emph{CVPR} (2023), pp.~4595--4605.
\newblock \href {https://doi.org/10.1109/CVPR52729.2023.00446}
  {\path{doi:10.1109/CVPR52729.2023.00446}}.

\bibitem[LCW{\etalchar{*}}24]{li2024loopgaussian}
\textsc{Li J., Cheng L., Wang Z., Mu T., He J.}:
\newblock {LoopGaussian}: Creating {3D} cinemagraph with multi-view images via
  {Eulerian} motion field.
\newblock In \emph{ACM MM} (2024), pp.~476--485.
\newblock \href {https://doi.org/10.1145/3664647.3681025}
  {\path{doi:10.1145/3664647.3681025}}.

\bibitem[LDZY24]{lin2023gaussianflow}
\textsc{Lin Y., Dai Z., Zhu S., Yao Y.}:
\newblock {Gaussian-Flow}: {4D} reconstruction with dynamic {3D Gaussian}
  particle.
\newblock In \emph{CVPR} (2024), pp.~21136--21145.
\newblock \href {https://doi.org/10.1109/CVPR52733.2024.01997}
  {\path{doi:10.1109/CVPR52733.2024.01997}}.

\bibitem[LFH15]{liao2015fastseamlessloops}
\textsc{Liao J., Finch M., Hoppe H.}:
\newblock Fast computation of seamless video loops.
\newblock \emph{ACM TOG 34}, 6 (2015).
\newblock \href {https://doi.org/10.1145/2816795.2818061}
  {\path{doi:10.1145/2816795.2818061}}.

\bibitem[LGB{\etalchar{*}}26]{luo2026growflow}
\textsc{Luo W., Goli L., Bahmani S., Taubner F., Tagliasacchi A., Lindell
  D.~B.}:
\newblock Grow with the flow: {4D} reconstruction of growing plants with
  {Gaussian} flow fields, 2026.
\newblock \href {http://arxiv.org/abs/2602.08958} {\path{arXiv:2602.08958}}.

\bibitem[LH96]{levoy1996lightfields}
\textsc{Levoy M., Hanrahan P.}:
\newblock Light field rendering.
\newblock In \emph{SIGGRAPH} (1996), p.~31–42.
\newblock \href {https://doi.org/10.1145/237170.237199}
  {\path{doi:10.1145/237170.237199}}.

\bibitem[LIM{\etalchar{*}}26]{le2026cinemagraphvideodiffusion}
\textsc{Le V.~T., Ito D., Mahapatra A., Mai L., Singh K.~K., Kulkarni K., Liu
  F., Fu Y., Yoon J.~S.}:
\newblock Controllable cinemagraph generation from a still image.
\newblock In \emph{EG Short} (2026), Eurographics Association.
\newblock \href {https://doi.org/10.2312/egs.20261010}
  {\path{doi:10.2312/egs.20261010}}.

\bibitem[LJH13]{liao2013automatedvideolooping}
\textsc{Liao Z., Joshi N., Hoppe H.}:
\newblock Automated video looping with progressive dynamism.
\newblock \emph{ACM TOG 32}, 4 (2013).
\newblock \href {https://doi.org/10.1145/2461912.2461950}
  {\path{doi:10.1145/2461912.2461950}}.

\bibitem[LKLR24]{luiten2023dynamic}
\textsc{Luiten J., Kopanas G., Leibe B., Ramanan D.}:
\newblock {Dynamic 3D Gaussians}: Tracking by persistent dynamic view
  synthesis.
\newblock In \emph{3DV} (2024), pp.~800--809.
\newblock \href {https://doi.org/10.1109/3DV62453.2024.00044}
  {\path{doi:10.1109/3DV62453.2024.00044}}.

\bibitem[MESK22]{mueller2022instant}
\textsc{M\"{u}ller T., Evans A., Schied C., Keller A.}:
\newblock Instant neural graphics primitives with a multiresolution hash
  encoding.
\newblock \emph{ACM TOG 41}, 4 (2022).
\newblock \href {https://doi.org/10.1145/3528223.3530127}
  {\path{doi:10.1145/3528223.3530127}}.

\bibitem[MK22]{mahapatra2021controllable}
\textsc{Mahapatra A., Kulkarni K.}:
\newblock Controllable animation of fluid elements in still images.
\newblock In \emph{CVPR} (2022), pp.~3657--3666.
\newblock \href {https://doi.org/10.1109/CVPR52688.2022.00365}
  {\path{doi:10.1109/CVPR52688.2022.00365}}.

\bibitem[MLLS23]{ma2023videoloop}
\textsc{Ma L., Li X., Liao J., Sander P.~V.}:
\newblock {3D} video loops from asynchronous input.
\newblock In \emph{CVPR} (2023), pp.~310--320.
\newblock \href {https://doi.org/10.1109/CVPR52729.2023.00038}
  {\path{doi:10.1109/CVPR52729.2023.00038}}.

\bibitem[MST{\etalchar{*}}20]{mildenhall2020nerf}
\textsc{Mildenhall B., Srinivasan P.~P., Tancik M., Barron J.~T., Ramamoorthi
  R., Ng R.}:
\newblock {NeRF}: Representing scenes as neural radiance fields for view
  synthesis.
\newblock In \emph{ECCV} (2020), pp.~405--421.
\newblock \href {https://doi.org/10.1007/978-3-030-58452-8_24}
  {\path{doi:10.1007/978-3-030-58452-8_24}}.

\bibitem[PCPMMN21]{pumarola2021dnerf}
\textsc{Pumarola A., Corona E., Pons-Moll G., Moreno-Noguer F.}:
\newblock {D-NeRF}: Neural radiance fields for dynamic scenes.
\newblock In \emph{CVPR} (2021), pp.~10313--10322.
\newblock \href {https://doi.org/10.1109/CVPR46437.2021.01018}
  {\path{doi:10.1109/CVPR46437.2021.01018}}.

\bibitem[PSB{\etalchar{*}}21]{park2021nerfies}
\textsc{Park K., Sinha U., Barron J.~T., Bouaziz S., Goldman D.~B., Seitz
  S.~M., Martin-Brualla R.}:
\newblock Nerfies: Deformable neural radiance fields.
\newblock In \emph{ICCV} (2021), pp.~5845--5854.
\newblock \href {https://doi.org/10.1109/ICCV48922.2021.00581}
  {\path{doi:10.1109/ICCV48922.2021.00581}}.

\bibitem[SF16]{schoenberger2016colmap1}
\textsc{Sch\"{o}nberger J.~L., Frahm J.-M.}:
\newblock Structure-from-motion revisited.
\newblock In \emph{CVPR} (2016), pp.~4104--4113.
\newblock \href {https://doi.org/10.1109/CVPR.2016.445}
  {\path{doi:10.1109/CVPR.2016.445}}.

\bibitem[SHK{\etalchar{*}}24]{shih2024ambientdynamics}
\textsc{Shih M.-L., Huang J.-B., Kim C., Shah R., Kopf J., Gao C.}:
\newblock Modeling ambient scene dynamics for free-view synthesis.
\newblock In \emph{SIGGRAPH} (2024).
\newblock \href {https://doi.org/10.1145/3641519.3657488}
  {\path{doi:10.1145/3641519.3657488}}.

\bibitem[SSC22]{sun2022dvgo}
\textsc{Sun C., Sun M., Chen H.}:
\newblock {Direct Voxel Grid Optimization}: Super-fast convergence for radiance
  fields reconstruction.
\newblock In \emph{CVPR} (2022), pp.~5449--5459.
\newblock \href {https://doi.org/10.1109/CVPR52688.2022.00538}
  {\path{doi:10.1109/CVPR52688.2022.00538}}.

\bibitem[SSSE00]{schodl2000vidtex}
\textsc{Sch\"{o}dl A., Szeliski R., Salesin D.~H., Essa I.}:
\newblock Video textures.
\newblock In \emph{SIGGRAPH} (USA, 2000), ACM Press/Addison-Wesley Publishing
  Co., pp.~489--498.
\newblock \href {https://doi.org/10.1145/344779.345012}
  {\path{doi:10.1145/344779.345012}}.

\bibitem[TD20]{teed2020raft}
\textsc{Teed Z., Deng J.}:
\newblock {RAFT}: Recurrent all-pairs field transforms for optical flow.
\newblock In \emph{ECCV} (2020), pp.~402--419.
\newblock \href {https://doi.org/10.1007/978-3-030-58536-5_24}
  {\path{doi:10.1007/978-3-030-58536-5_24}}.

\bibitem[THM{\etalchar{*}}26]{tzathas2026oscilatorymotion}
\textsc{Tzathas P., Hu J., Meuleman A., Cordonnier G., Drettakis G.}:
\newblock Adaptive spatio-temporal {3D Gaussian} splatting for scenes with
  oscillatory motion.
\newblock \emph{Comput. Graph. Forum} (2026), e70410.
\newblock \href {https://doi.org/10.1111/cgf.70410}
  {\path{doi:10.1111/cgf.70410}}.

\bibitem[TIF24]{torralba2024foundcv}
\textsc{Torralba A., Isola P., Freeman W.~T.}:
\newblock \emph{Foundations of Computer Vision}.
\newblock Adaptive Computation and Machine Learning series. MIT Press,
  Cambridge, MA, 2024.
\newblock URL:
  \url{https://mitpress.mit.edu/9780262048972/foundations-of-computer-vision/}.

\bibitem[TPSK11]{tompkin2011towardsmomentimagery}
\textsc{Tompkin J., Pece F., Subr K., Kautz J.}:
\newblock Towards moment imagery: Automatic cinemagraphs.
\newblock In \emph{CVMP} (2011), pp.~87--93.
\newblock \href {https://doi.org/10.1109/CVMP.2011.16}
  {\path{doi:10.1109/CVMP.2011.16}}.

\bibitem[WYF{\etalchar{*}}24]{wu20244dgs}
\textsc{Wu G., Yi T., Fang J., Xie L., Zhang X., Wei W., Liu W., Tian Q., Wang
  X.}:
\newblock {4D Gaussian} splatting for real-time dynamic scene rendering.
\newblock In \emph{CVPR} (2024), pp.~20310--20320.
\newblock \href {https://doi.org/10.1109/CVPR52733.2024.01920}
  {\path{doi:10.1109/CVPR52733.2024.01920}}.

\bibitem[XZQ{\etalchar{*}}24]{xie2024physgaussian}
\textsc{Xie T., Zong Z., Qiu Y., Li X., Feng Y., Yang Y., Jiang C.}:
\newblock {PhysGaussian}: Physics-integrated {3D Gaussians} for generative
  dynamics.
\newblock In \emph{CVPR} (2024), pp.~4389--4398.
\newblock \href {https://doi.org/10.1109/CVPR52733.2024.00420}
  {\path{doi:10.1109/CVPR52733.2024.00420}}.

\bibitem[YYPZ24]{yang20244dgaussians}
\textsc{Yang Z., Yang H., Pan Z., Zhang L.}:
\newblock Real-time photorealistic dynamic scene representation and rendering
  with {4D Gaussian} splatting.
\newblock In \emph{ICLR} (2024).
\newblock URL: \url{https://openreview.net/forum?id=WhgB5sispV}.

\bibitem[ZPvBG01]{zwicker2001ewasplat}
\textsc{Zwicker M., Pfister H., van Baar J., Gross M.}:
\newblock {EWA} volume splatting.
\newblock In \emph{IEEE Vis.} (2001), pp.~29--38.
\newblock \href {https://doi.org/10.1109/VISUAL.2001.964490}
  {\path{doi:10.1109/VISUAL.2001.964490}}.

\end{thebibliography}

\appendix
\section{Dense Flow Computation}
\label{app:dense}
Following Mahapatra and Kulkarni~\cite{mahapatra2021controllable}, we derive dense flow maps $\mathcal{F}^{\textit{dense}}_{1,\cdots,N}$ from sparse flow maps $\mathcal{F}^{\textit{sparse}}_{1,\cdots,N}$ by spreading their sparse flow with weighted exponential pixel-distance functions defined as
\begin{equation}
    D(p_1,\,p_2) = e^{-\frac{\lVert p_1-p_2 \rVert_2^2}{\sigma^2}}\,.
\end{equation}
Here the L2 Euclidean distance is used in conjunction with a constant $\sigma = 0.075H$.
For each pixel $x$, sparse flow $\mathcal{F}^{\textit{sparse}}_n(p)$ is weighted by $D(x,\,p)$ based on hint position $p$.
Pixel values are summed up for all sparse hints and normalized to define dense flow maps:
\begin{equation}
    \mathcal{F}^{\textit{dense}}_n(x)=
    \begin{cases}
    \frac{
        \sum_{i=1}^{|\mathcal{H}_n|} D(x,\,p_i) \cdot \mathcal{F}^{\textit{sparse}}_n(p_i)
    }{
        \sum_{i=1}^{|\mathcal{H}_n|} D(x,\,p_i)
    }
    , & \text{if } x\in\mathcal{S}_n \\
    0, & \text{otherwise}
    \end{cases} \,,
\end{equation}
with $p_i$ the $i$-th hint position in $\mathcal{H}_n$.
Note that dense flow is only defined for pixels within the segmentation mask of the view $\mathcal{S}_n$.

\section{3D Motion Field Creation}
\label{app:motion_field_creation}
\cref{algo:fuse} outlines our approach for 3D motion field fusion and refinement.
\begin{algorithm}[htb]
\caption{\scshape Fuse-Motion-Field}
\label{algo:fuse}
\begin{algorithmic}[1]
    \REQUIRE $\mathcal{F}_{1,\cdots,N}$, $\mathcal{D}_{1,\cdots,N}$, $\mathcal{S}_{1,\cdots,N}$ 
    \ENSURE $\mathcal{M}$
    \FOR[dynamic region fusion]{\textbf{each} $n\in\{1,\cdots,N\}$}
        \FOR{\textbf{each} $(\mathbf{c}, v_n, w_n) \in\text{\scshape GroupDynamicPixels}(\mathcal{S}_n,\, \mathcal{F}_n,\, \mathcal{D}_n)$}
            \STATE $\mathbf{A}_{\mathbf{c}} \mathrel{+}= J^\top_{n,\,\mathbf{c}} J_{n,\,\mathbf{c}}$
            \STATE $\mathbf{b}_{\mathbf{c}} \mathrel{+}= J^\top_{n,\,\mathbf{c}} v_n$ \COMMENT{accumulate $\mathbf{A}$ and $\mathbf{b}$}
            \STATE $w^{\text{dyn}}_{\mathbf{c}} \mathrel{+}= w_n$ \COMMENT{track dynamic pixel hits}
            \STATE $w^{\text{view}}_{\mathbf{c}} \mathrel{+}= 1$ \COMMENT{track view hits}
          \ENDFOR
    \ENDFOR
    \STATE $\mathcal{M} \gets \text{\scshape Solve}(C,\,\mathbf{A},\, \mathbf{b})$
    \FOR[static region refinement]{\textbf{each} $n\in\{1,\cdots,N\}$}
        \FOR{\textbf{each} $(\mathbf{c},w_n)\in\text{\scshape GroupStaticPixels}(\mathcal{S}_n,\, \mathcal{D}_n)$}
            \STATE $w^{\text{stat}}_{\mathbf{c}} \mathrel{+}= w_n$ \COMMENT{track static pixel hits}
          \ENDFOR
    \ENDFOR
    \STATE $\mathcal{M} \gets \text{\scshape Refine}(C,\, w^{\text{stat}},\, w^{\text{dyn}},\, w^{\text{view}})$
\end{algorithmic}
\end{algorithm}

\section{Motion Field Properties}
\label{app:motion_field_prop}

In \cref{tab:motion_field}, we further analyze per scene motion field properties, where grid size is clearly reduced by our refinement and the density of the refined grid does not exceed $5\%$, enabling memory-efficient, sparse storage.
We set the initial number of voxels individually per scene to constrain VRAM during fusion below 16GB.
We find that higher resolution grids do not largely benefit animation quality for our tested scenes.

\begin{table}[b]
\caption[Properties of motion fields]{%
Properties of per scene motion fields including initial and refined number of voxels $|C|$, density $\rho$ of the refined grid, average view hits per voxel $w^{\text{view}}$ and view hit threshold $r^{\text{view}}$ used.
}\label{tab:motion_field}
\vspace{1em}
\centering
\setlength\tabcolsep{8.0pt}
\scriptsize
\begin{tabular}{@{}lccccc}
\toprule
Scene & initial $|C|$ & refined $|C|$ & refined $\rho$ & average $w^{\text{view}}$ & $r^{\text{view}}$ \\
\midrule
Creek & 75\,M & 14\,M & 2.7\,\% & 9 & 5 \\
Frogfall & 50\,M & 2.6\,M & 0.1\,\% & 37 & 8 \\
Silverfalls & 70\,M & 0.3\,M & 4.7\,\% & 12 & 8 \\
Smallfall & 50\,M & 1.4\,M & 3.6\,\% & 16 & 7 \\
Town & 60\,M & 2.0\,M & 4.1\,\% & 19 & 6 \\
Waterfall & 60\,M & 34\,M & 0.8\,\% & 23 & 8 \\
Weir & 70\,M & 11\,M & 2.2\,\% & 22 & 12 \\
\bottomrule
\end{tabular}
\end{table}

\begin{figure*}[t]
\centering
\includegraphics[width=\linewidth]{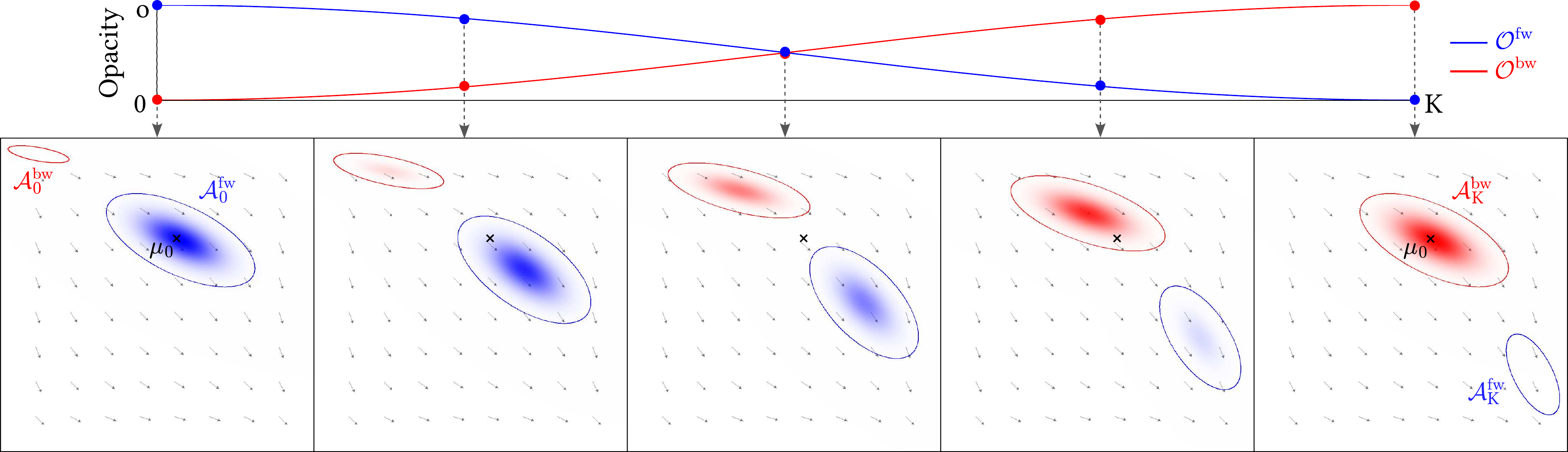}
\caption[symmetric animation technique]{
Illustration of our symmetric animation technique for one splatted Gaussian. Forward instance $\mathcal{A}^{\text{fw}}$ (blue) and backward instance $\mathcal{A}^{\text{bw}}$ (red) are illustrated at different keyframes.
Additionally, the Gaussian contour is drawn independent of Gaussian opacity, where $\mathcal{O}^{\text{fw}}$ and $\mathcal{O}^{\text{bw}}$ are plotted.
The Gaussian primitive is deformed by the underlying motion field (see \cref{sec:animation_ut}).
}
\label{fig:animation}
\end{figure*}

\section{Animation Illustration and Ablations}
\label{app:animation}

\begin{figure}
\centering
\includegraphics[width=0.94\linewidth]{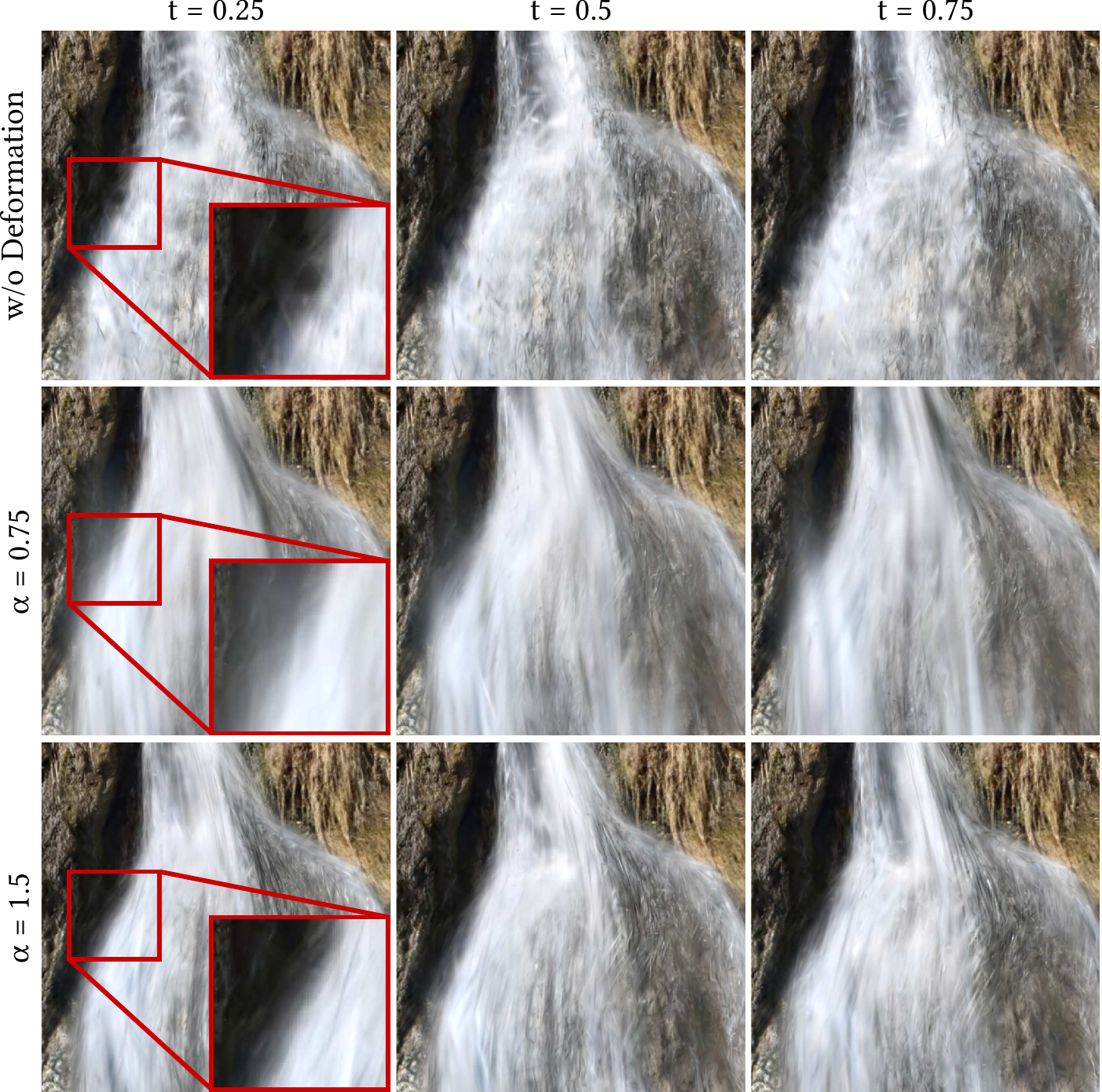}
\caption[ablation of deformation]{
Ablation of Gaussian transformation. Only displacing Gaussian means with constant covariance creates noisy artifacts (top row).
A small $\alpha$ (see \cref{eq:sigma_points}) results in blurry unrealistic deformations (middle row), while a larger $\alpha$ creates convincing transformations following fluid motion.
}
\label{fig:ablation_deform}
\end{figure}

In \cref{fig:animation} we illustrate a 2D example of the Gaussian deformation and symmetric animation technique.
In \cref{fig:ablation_deform}, we evaluate the Unscented Transform-based Gaussian transformation technique.
In the top row, only the Gaussian mean is displaced, while covariance remains constant.
As a result, elongated Gaussians, not pointing in the direction of fluid motion, create noisy artifacts.
We further alter $\alpha$ (see \cref{eq:sigma_points}) controlling the spread of sigma points around the mean.
We find that smaller values produce unrealistic deformations, further introducing blur, while larger values result in moderate deformations following fluid motion.

To achieve high-quality results in our qualitative comparison, we employ $K=60$ animation keyframes. However, the use of fewer keyframes can significantly reduce memory consumption.
In order to create the desired effect of flowing fluid, Gaussian particles need to travel a sufficient distance through the scene.
Since decreasing $K$ results in shorter trajectories, a larger motion regulation $r^\mathcal{M}$, increasing particle displacement per keyframe, can accommodate the effect.
If $r^\mathcal{M}$ is too large, trajectories might diverge and degenerate, but we find that decreasing to $K=8$ and increasing $r^\mathcal{M}$ accordingly still produces visually plausible animations and maintains smoothness due to keyframe interpolation.

\section{Positional Offsets}
\label{app:pos_offset}
\begin{figure}
\centering
\includegraphics[width=0.94\linewidth]{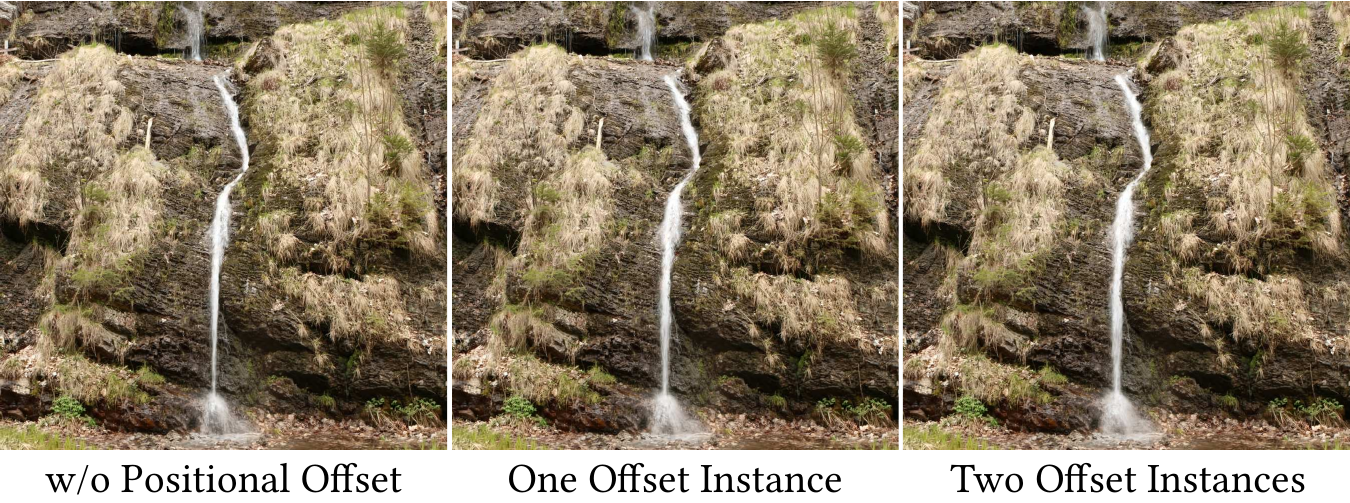}
\caption[ablation of positional offset]{
We demonstrate the use of additional positional offset instances. While without positional offset (left) the waterfall is sparsely reconstructed, additional instances (right) create more voluminous and denser fluid regions.
}
\label{fig:ablation_pos_off}
\end{figure}

Additional transformation sequences can be defined by utilizing the symmetric animation technique (see \cref{sec:animation_symmetric}).
Adding a random per Gaussian positional offset $\mu^{\text{off}} \in \mathbb{R}^3$ to $\mu$ before transforming a Gaussian $g$ forward and backward creates new trajectories similar to $\mathcal{A}^\text{fw}$ and $\mathcal{A}^\text{bw}$.
We sample $\mu^{\text{off}}$ uniformly from a spherical shell with restricted maximum length.
The transformations are blended with unaltered $\mathcal{O}^\text{fw}$ and $\mathcal{O}^\text{bw}$ to enable the same looping behavior for the new instances.
Furthermore, the per Gaussian random animation state offset $t^{\text{off}}$ is applied.
This procedure can be repeated to generate multiple offset transformation trajectories of the same $g$, which is useful for enhancing scene density of sparsely reconstructed fluid regions and enables customization of fluid volume.
We find that more than two additional instances increase density to the point where animation detail is lost.
With increasing offset extent, the scene appearance degenerates and introduces artifacts, observable in \cref{fig:ablation_pos_off}.

\end{document}